\documentclass[sigconf]{acmart}

\usepackage{graphicx}
\usepackage{subcaption}
\usepackage{booktabs}
\usepackage{multirow}
\usepackage{tcolorbox}
\usepackage{xcolor}
\usepackage{makecell} 
\usepackage{balance}
\definecolor{lightgray}{gray}{0.9}
\usepackage[skip=3pt]{caption}  

\AtBeginDocument{%
  }

\setcopyright{acmlicensed}
\copyrightyear{2018}
\acmYear{2018}
\acmDOI{XXXXXXX.XXXXXXX}
\acmConference[Conference acronym 'XX]{Make sure to enter the correct
  conference title from your rights confirmation email}{June 03--05,
  2018}{Woodstock, NY}
\acmISBN{978-1-4503-XXXX-X/2018/06}

\begin{document}

\title{\textsc{Dynamic-KGQA}: A Scalable Framework for Generating Adaptive Question Answering Datasets}

\author{Preetam Prabhu Srikar Dammu}
\affiliation{
    \institution{University of Washington}
    \city{Seattle}
    \state{Washington}
    \country{USA}
}
\email{preetams@uw.edu}

\author{Himanshu Naidu}
\affiliation{
    \institution{University of Washington}
    \city{Seattle}
    \state{Washington}
    \country{USA}
}
\email{hnaidu36@uw.edu}

\author{Chirag Shah}
\affiliation{
    \institution{University of Washington}
    \city{Seattle}
    \state{Washington}
    \country{USA}
}
\email{chirags@uw.edu}

\renewcommand{\shortauthors}{Dammu et al.}

\begin{abstract}

As question answering (QA) systems advance alongside the rapid evolution of foundation models, the need for robust, adaptable, and large-scale evaluation benchmarks becomes increasingly critical. Traditional QA benchmarks are often static and publicly available, making them susceptible to data contamination and memorization by large language models (LLMs). Consequently, static benchmarks may overestimate model generalization and hinder a reliable assessment of real-world performance.
In this work, we introduce \textsc{Dynamic-KGQA}, a scalable framework for generating adaptive QA datasets from knowledge graphs (KGs), designed to mitigate memorization risks while maintaining statistical consistency across iterations. Unlike fixed benchmarks, \textsc{Dynamic-KGQA} generates a new dataset variant on every run while preserving the underlying distribution, enabling fair and reproducible evaluations. Furthermore, our framework provides fine-grained control over dataset characteristics, supporting domain-specific and topic-focused QA dataset generation. Additionally, \textsc{Dynamic-KGQA} produces compact, semantically coherent subgraphs that facilitate both training and evaluation of KGQA models, enhancing their ability to leverage structured knowledge effectively.
To align with existing evaluation protocols, we also provide static large-scale train/test/validation splits, ensuring comparability with prior methods. By introducing a dynamic, customizable benchmarking paradigm, \textsc{Dynamic-KGQA} enables a more rigorous and adaptable evaluation of QA systems.

\end{abstract}



\maketitle

\section{Introduction}


Generative models have transformed question-answering (QA) systems by enabling users to pose complex queries naturally, without requiring precise formulations or extensive manual searching. They are capable of interpreting nuanced questions, performing multi-hop reasoning across diverse sources, and effectively disambiguating user intent \cite{lewis2020retrieval,luo2024reasoning,chen2024planongraph,sun2024thinkongraph,li2024survey}. However, these models are prone to hallucinations, especially when handling out-of-distribution queries or those underrepresented in training data \cite{huang2023survey,li2024survey}.

Research indicates that integrating external knowledge can enhance the quality, reliability, and factual accuracy of answers \cite{lewis2020retrieval,luo2024reasoning,chen2024planongraph,sun2024thinkongraph}. For queries requiring complex reasoning, knowledge graph question-answering (KGQA) methods show promising results, as KGs provide explicit, traceable information and support multi-hop reasoning \cite{zhang2022subgraph,ding2024enhancing,jiang2022unikgqa,luo2024reasoning,chen2024planongraph,sun2024thinkongraph}. Recent agent-based KGQA methods further improve adaptability by enabling dynamic exploration, self-correction, and efficient knowledge retrieval, even outperforming larger models in certain tasks \cite{sun2024thinkongraph,chen2024planongraph,luo2024reasoning}.

Despite these advancements, the benchmarks currently used for evaluating newer-generation QA systems remain predominantly static, consisting of fixed datasets that are publicly accessible. Several studies show significant evidence that many widely used benchmarks have already been contaminated, rendering them unreliable \cite{zhou2023don,yang2023rethinking,Zhu2023DyValDE,Zhu2024DyVal2D,golchin2023data,li2023open}. \citet{zhou2023don} refer to this issue as benchmark leakage, a growing concern as large foundation models are trained on web-scale datasets that encompass vast portions of publicly available internet data \cite{yang2023rethinking,Zhu2023DyValDE,vukov2023ouroboros}. Studies indicate that even simple paraphrasing can degrade performance, emphasizing the brittle nature of these systems and their reliance on memorization \cite{Zhu2023DyValDE}.

To overcome these limitations, dynamic benchmarks have emerged as a promising alternative \cite{kiela2021dynabench,Zhu2024DyVal2D,Zhu2023DyValDE,ma2021dynaboard,zhang2024darg}. Unlike their static counterparts, dynamic benchmarks are designed to resist memorization and provide a more accurate assessment of adaptive, context-aware performance in evolving scenarios \citep{yao2024tau, Zhu2024DyVal2D, Zhu2023DyValDE, rawles2024androidworld}. This shift is particularly important for ensuring that reported evaluations reflect genuine advancements in capabilities rather than superficial performance gains.

Furthermore, static benchmark datasets do not provide control over the domain or topic of the content, despite the need for domain-specific QA datasets. For instance, MedQA \cite{jin2021disease} is a QA dataset specifically designed for medical queries. However, when suitable datasets are not readily available, creating them is a non-trivial task, hindering research progress. In contrast, dynamic benchmark frameworks help overcome this challenge by enabling the generation of domain-specific datasets in a flexible and scalable manner.

In this work, we introduce \textsc{Dynamic-KGQA}, a novel benchmarking framework for KGQA that generates dynamic samples, reducing the risk of data contamination. \textsc{Dynamic-KGQA} exhibits several desirable characteristics, as outlined below:

\begin{enumerate} 
    \item Ensures statistical consistency across different runs while generating dynamic QA samples.
    \item Constructs compact, thematic subgraphs for each QA pair, enabling controlled testing and supporting KGQA research. 
    \item Provides static train/test/validation splits to facilitate comparison studies with earlier benchmarks. 
    \item Enables controlled dataset generation, allowing researchers to specify the domain or topic of interest.
\end{enumerate}

We make the dataset available on Hugging Face\footnote{\url{https://huggingface.co/datasets/preetam7/dynamic_kgqa}}, the source code on GitHub\footnote{\url{https://github.com/PreetamDammu/dynamicKGQA}}, and additional resources on our website\footnote{\url{https://www.preetamdammu.com/dynamic-kgqa-web/}} to support research efforts in the dynamic evaluation of KGQA systems.

\begin{figure*}[h]
    \centering
    \includegraphics[width=0.95\linewidth]{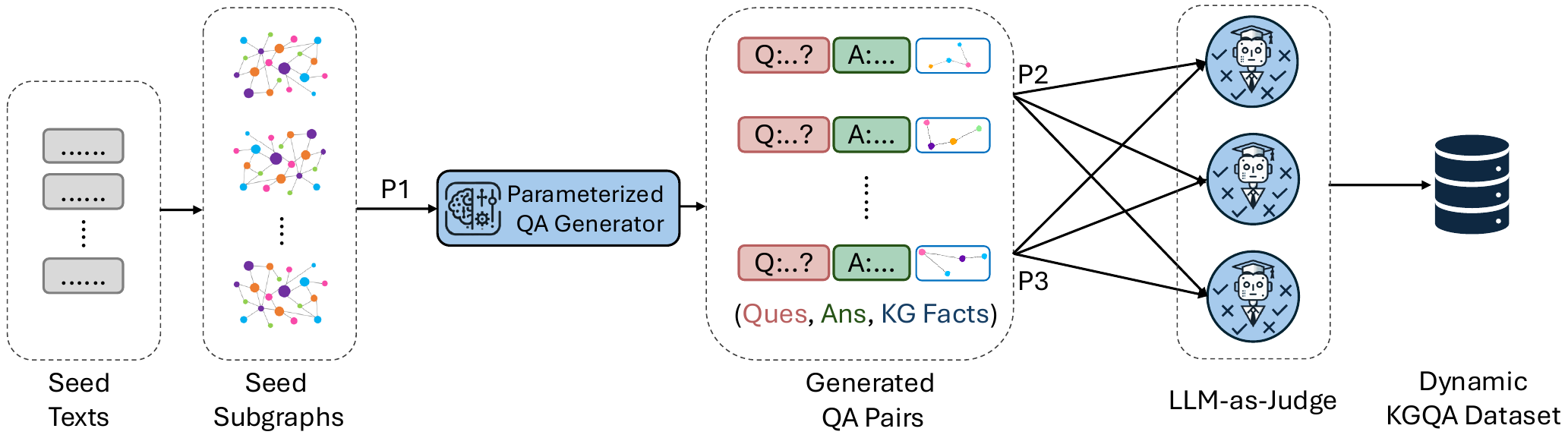}
    \caption{Overview of \textsc{Dynamic-KGQA}. The framework generates dynamic QA samples while ensuring statistical consistency across runs. Given seed texts, subgraphs are extracted from the KG to ground QA generation. The generated QA pairs are assessed for structural validity, redundancy, and correctness. See prompts P1–P3 in \ref{app:prompts} for details.}
    \label{fig:pipeline}
\end{figure*}

\section{Related Work}
\label{sec:relatedWork}
\vspace{2mm}
\noindent{\textbf{\textit{KGQA Methods}}}. 
Modern KGQA approaches increasingly focus on balancing structural reasoning with efficient retrieval \cite{zhang2022subgraph,saxena2020improving,ding2024enhancing}. Early semantic parsing methods faced scalability challenges due to dependency on annotated logical forms or restricted predicates \cite{lan2020query,berant2014semantic,bao2016constraint,das2021case}. Subsequent embedding-based systems improved flexibility but struggled with noise from irrelevant entities in unconstrained search spaces \cite{saxena2020improving,miller2016key,zhang2018variational}.
More recent work emphasizes dynamic subgraph extraction to isolate question-relevant evidence while preserving graph topology \cite{zhang2022subgraph,ding2024enhancing,jiang2022unikgqa,sun2019pullnet,sun2018open,saxena2020improving,jiang2023reasoninglm,wu2023retrieve,zhang2018variational}. However, the absence of benchmark subgraph annotations complicates isolating retrieval-vs-reasoning bottlenecks, a concern we address through the proposed framework.

As LLMs' reasoning and tool-calling capabilities improve, an increasing number of methods leverage LLMs for question answering over KGs, reporting higher scores than previous approaches \cite{sun2024thinkongraph,chen2024planongraph,luo2024reasoning}. These methods iteratively search or navigate the KG to identify the most promising reasoning paths for obtaining the correct answer. Such approaches would benefit from KGQA benchmarks optimized for automated reasoning -- free from taxonomic loops and featuring human-readable paths and entity names -- characteristics often missing in existing KGQA datasets. We address these gaps in our work.

\vspace{2mm}
\noindent{\textbf{\textit{KGQA Datasets}}}. 
A persistent challenge in KGQA research stems from the widespread adoption of Freebase as the underlying knowledge source for major benchmarks. 
Most foundational KGQA datasets – including WebQuestions \cite{berant2013semantic}, WebQSP \cite{yih2016value}, GrailQA \cite{gu2021beyond}, and CWQ \cite{talmor2018web} – derive from Freebase \cite{bollacker2008freebase}, a KG discontinued in 2015 \cite{pellissier2016freebase}. This restricts their relevance in real-world applications where up-to-date knowledge is essential. For instance, WebQSP\cite{yih2016value}, designed to evaluate semantic parsing, inherits Freebase’s static schema, making it incompatible with modern KGs. Models trained on Freebase-based datasets often fail to generalize to live knowledge bases like Wikidata due to schema divergence and temporal misalignment.

Newer benchmarks like QALD-9 \cite{ngomo20189th}, QALD10-en \cite{usbeck2024qald}, LC-QuAD \cite{dubey2019lc}, and LC-QuAD 2.0 \cite{trivedi2017lc} have transitioned towards DBpedia \cite{lehmann2015dbpedia} and Wikidata \cite{vrandevcic2014wikidata} as the base KGs. While these KGs are better maintained and more current, they are not particularly suitable for automated reasoning, as their collaborative nature introduces convoluted schemas, taxonomic loops, and other redundancy issues. The presence of such structural issues and noise may make them less ideal for LLM-based KGQA approaches, which rely on active exploration of the graph, and each step adds to costs, as LLM inference calls may not necessarily be cost-efficient. We address these shortcomings by using YAGO 4.5 \cite{suchanek2024yago}, a recent and cleaner KG with a schema compatible with automated reasoning, as the base KG for \textsc{Dynamic-KGQA}.

\vspace{2mm}
\noindent{\textbf{\textit{Dynamic Evaluation}}}. 
Dynamic evaluation has emerged as a more robust and scalable alternative to static benchmarks, which are increasingly susceptible to data contamination in LLM-based methods. Early attempts at dynamic benchmarking rely on crowdsourcing for data collection \cite{ma2021dynaboard,kiela2021dynabench}, making them costly and difficult to scale. More recent approaches leverage graph-based methods to generate evaluation samples \cite{Zhu2023DyValDE,Zhu2024DyVal2D,zhang2024darg}, offering advantages such as controllable complexity and adaptability to evolving requirements -- an essential feature given the rapid advancements in foundation models. While these approaches have proven effective in reasoning tasks \cite{Zhu2023DyValDE,Zhu2024DyVal2D,zhang2024darg}, dynamic benchmarking for KGQA remains largely unexplored. We address this gap with \textsc{Dynamic-KGQA}, extending dynamic evaluation techniques to QA systems.

\vspace{2mm}
\noindent{\textbf{\textit{LLM-as-a-Judge}}}. 
The use of LLMs as evaluators, termed ``LLM-as-a-Judge,'' is gaining traction as a promising paradigm for automating complex assessment tasks. \citet{zheng2024judging} conduct a large-scale crowdsourced study to demonstrate that models like GPT-4 can achieve agreement levels comparable to human evaluators. Some studies even suggest that LLMs can outperform crowd-workers on certain NLP tasks \cite{gilardi2023chatgpt}. While LLMs offer a scalable alternative to manual evaluation, concerns around reliability and bias remain, particularly for subjective assessments \cite{huang2023chatgpt,gu2024survey,zheng2024judging}. However, for inherently objective tasks like KGQA, which share similarities with fact verification and natural language inference -- areas where LLMs have demonstrated strong performance \cite{guan2023language,raffel2020exploring} -- LLM-as-a-Judge serves as a well-suited evaluation framework.

\section{Methodology}
\label{sec:method}

The framework is designed with two primary objectives: (1) to generate dynamic samples, reducing the risk of memorization or data contamination, and (2) to enforce consistency across different runs by anchoring generation to a controlled subgraph. 
The subsequent sections describe the key components of our approach. In \S \ref{subsec:constructingSubG}, we outline the procedure for extracting a subgraph from the KG, ensuring relevance and compactness. We then detail the QA generation process in \S \ref{subsec:generatingQA}, followed by techniques for introducing controlled variability into the generated QA pairs in \S \ref{subsec:randomizingQA}. To ensure that generated answers and reasoning paths remain grounded in the KG, we introduce a verification step in \S \ref{subsec:verifyQA}. Finally, in \S \ref{subsec:annotationLLM}, we describe the use of LLM-as-a-Judge for assessing the quality and correctness of the generated QA pairs.

\subsection{Constructing Seed Subgraphs}
\label{subsec:constructingSubG}

Let $\mathcal{G} = (\mathcal{V}, \mathcal{P}, \mathcal{T})$ represent a knowledge graph (KG), where:
\begin{itemize}
    \item $\mathcal{V}$ is the set of entities (nodes),
    \item $\mathcal{P}$ is the set of predicates (relationships),
    \item $\mathcal{T} \subseteq \mathcal{V} \times \mathcal{P} \times \mathcal{V}$ is the set of RDF triples of the form $(h, p, t)$, where $h$ (head) and $t$ (tail) are entities in $\mathcal{V}$, and $p$ (predicate) is a relation in $\mathcal{P}$.
\end{itemize}

The goal is to extract a subgraph $\mathcal{G}' = (\mathcal{V}', \mathcal{P}', \mathcal{T}')$, where:
\begin{enumerate}
    \item $\mathcal{G}'$ is thematically coherent and focuses on a specific topic,
    \item $\mathcal{G}'$ retains only relevant entities and relationships,
    \item $\mathcal{G}'$ is compact, suitable for generating QA pairs.
\end{enumerate}


\subsubsection{Selecting Seed Texts}
To construct semantically coherent subgraphs, the first step is to identify a set of nodes in the KG that are relevant to a specific topic. This is done by leveraging seed texts, which serve as an initial source for extracting entities and defining the subgraph’s scope. The choice of seed text directly determines the topic of the extracted subgraph.

\begin{figure*}[t]
    \centering
    \begin{subfigure}[b]{0.24\textwidth}
        \centering
        \includegraphics[width=\textwidth]{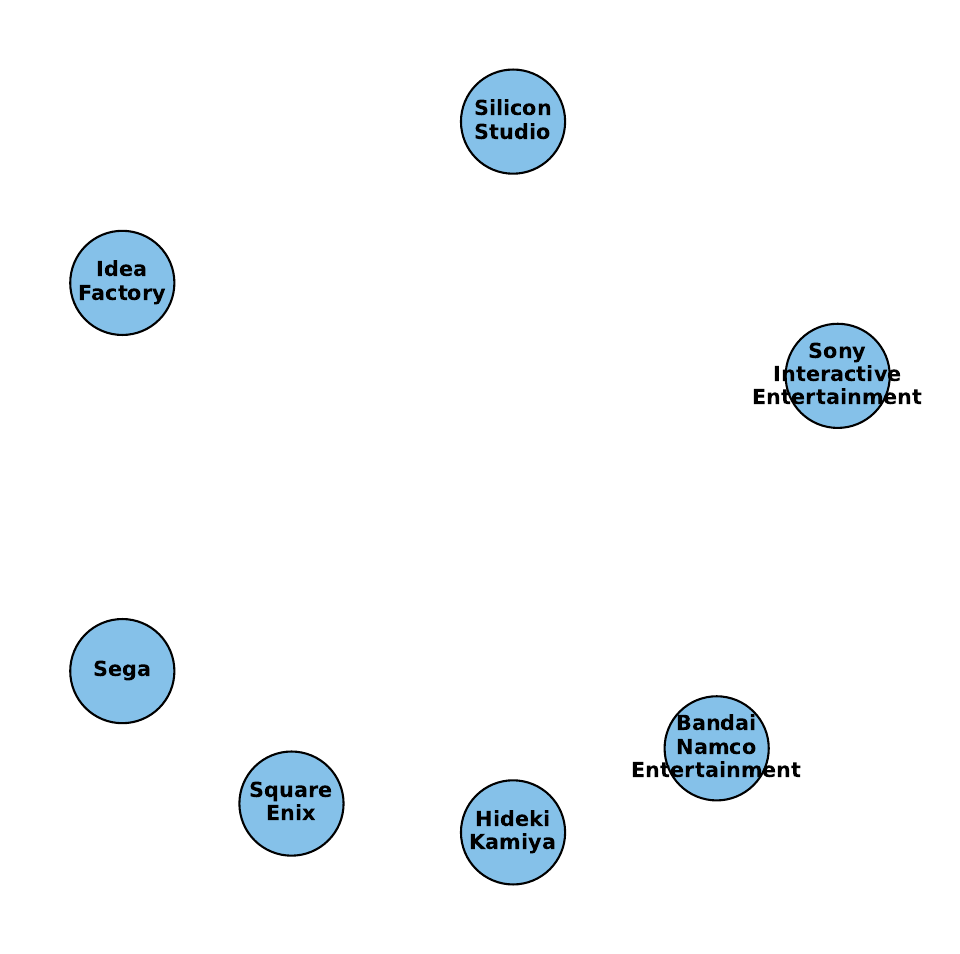}
        \caption{Seed entities $\mathcal{S}$.}
    \end{subfigure}
    \hfill
    \begin{subfigure}[b]{0.24\textwidth}
        \centering
        \includegraphics[width=\textwidth]{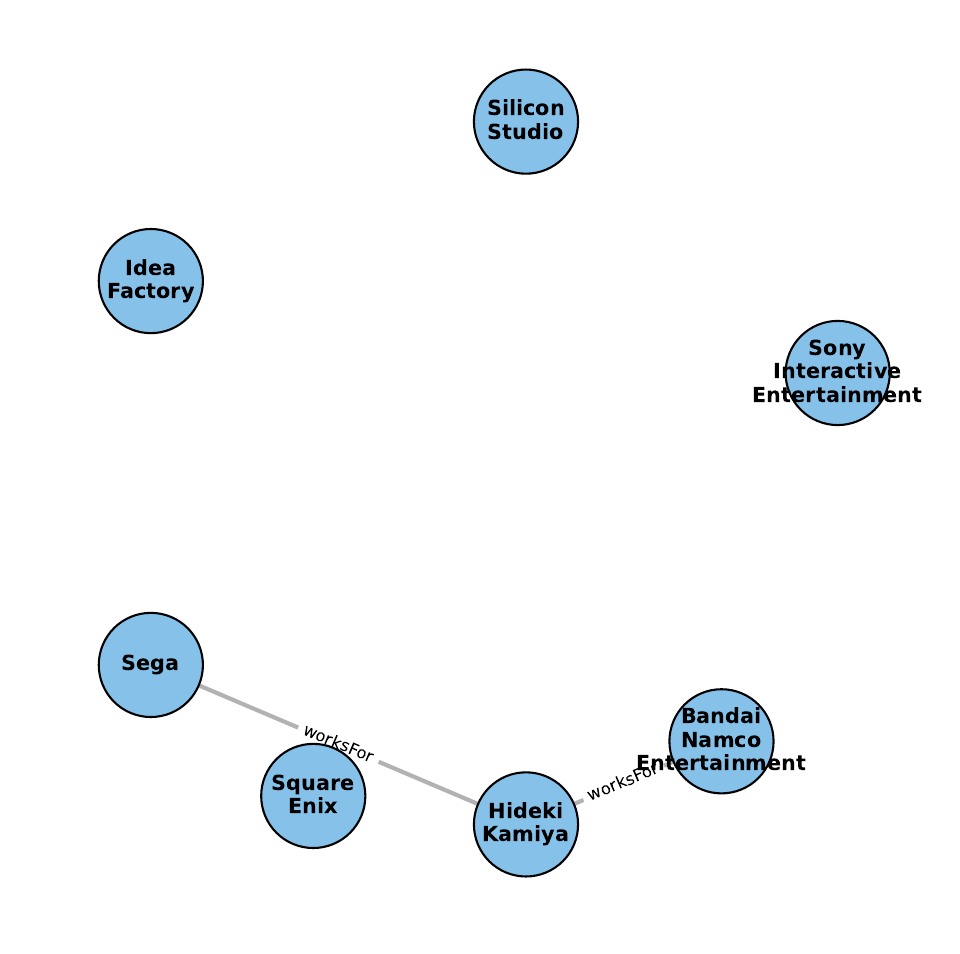}
        \caption{Direct seed links.}
    \end{subfigure}
    \hfill
    \begin{subfigure}[b]{0.24\textwidth}
        \centering
        \includegraphics[width=\textwidth]{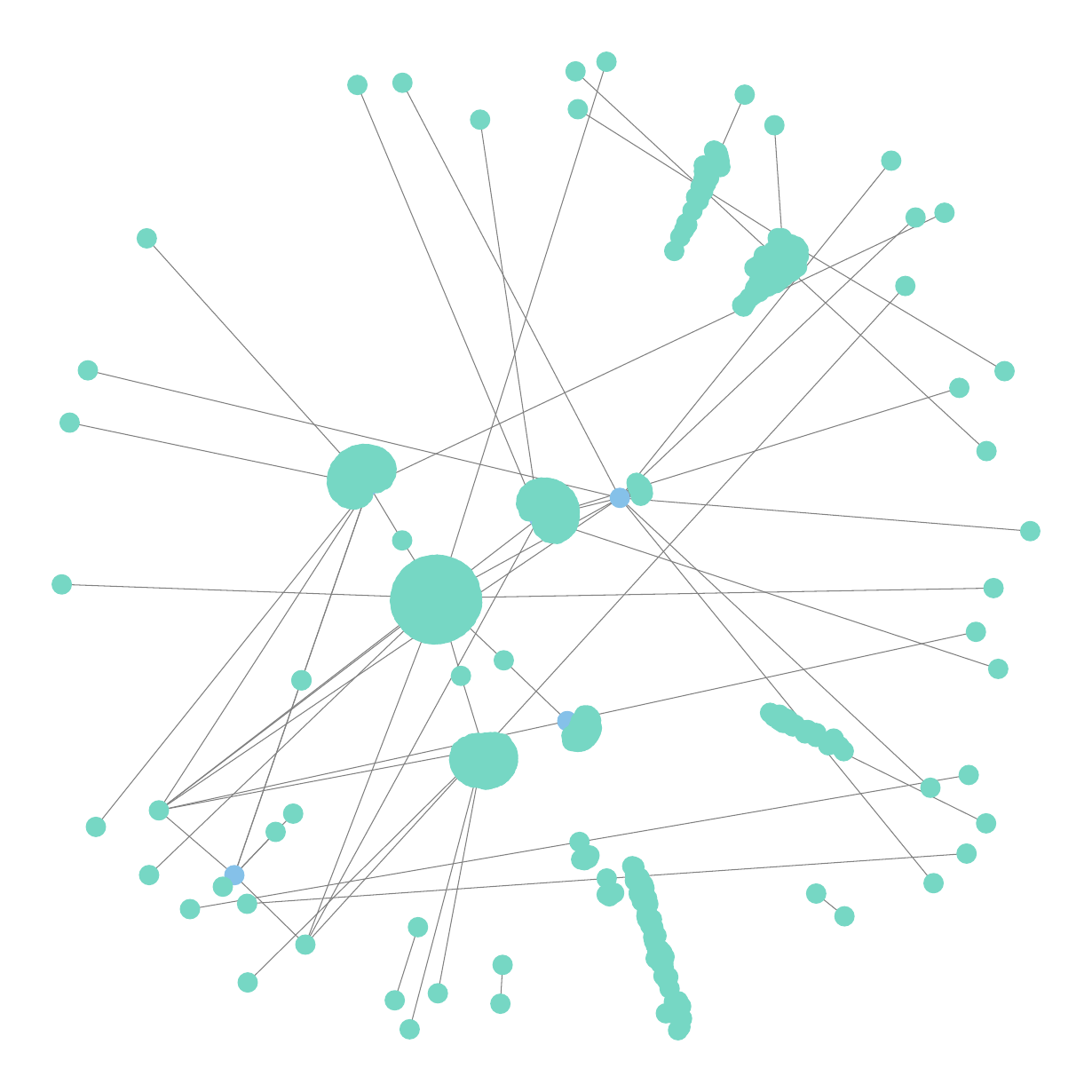}
        \caption{Expanded neighborhood.}
    \end{subfigure}
    \hfill
    \begin{subfigure}[b]{0.24\textwidth}
        \centering
        \includegraphics[width=\textwidth]{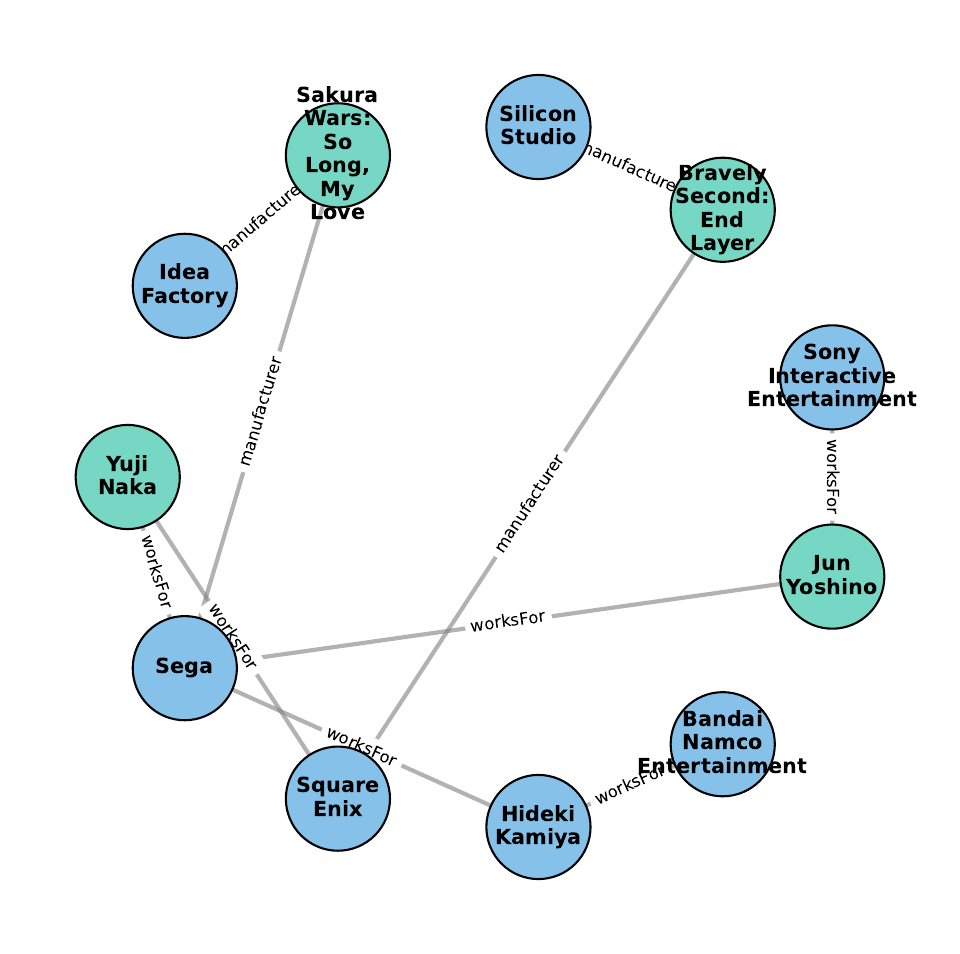}
        \caption{Steiner tree subgraph.}
    \end{subfigure}
    \caption{Process of constructing a structured subgraph from seed entities. (a) Initial seed entities $\mathcal{S}$ identified from the seed text. (b) Direct edges among $\mathcal{S}$ reveal sparse connectivity. (c) Neighborhood expansion via one-hop triples $\mathcal{T}_{\text{one-hop}}$ yields a denser graph. (d) Steiner tree extraction refines the graph to a minimal, coherent structure while filtering noise. \label{fig:graphConst}}
\end{figure*}

Selecting seed texts from a particular domain constrains the dataset to a specific subject area. A diverse selection results in a more general dataset, while specialized topics, such as historical wars, yield subgraphs and QA pairs focused on warfare. Additionally, subgraph size depends on seed text length -- longer texts with more entities produce larger subgraphs, whereas shorter texts lead to more compact ones. The core steps in constructing a structured subgraph from seed entities are visualized in Figure \ref{fig:graphConst}.

\subsubsection{Identifying Seed Entities}
Starting with a \emph{seed text}, we extract all \textit{entities} mentioned in it that exist in the KG. The process involves:
\begin{enumerate}
    \item Detecting entities $\mathcal{E}$ in \emph{seed text}.
    \item Retaining entities that exist in $\mathcal{V}$.
    \begin{equation}
        \mathcal{S} = \{ v \in \mathcal{E} \cap \mathcal{V} \}.
    \end{equation}
\end{enumerate}
For example, given a passage about Harry Potter, entities such as ``Harry Potter,'' ``Hogwarts,'' and ``Albus Dumbledore'' could be identified and linked to corresponding nodes in $\mathcal{V}$.

\subsubsection{Expanding the Neighborhood}

For each $v \in \mathcal{S}$, include all RDF triples involving $v$ as a subject or object:
\begin{equation}
    \mathcal{T}_{\text{one-hop}}(v) = \{ (h, p, t) \in \mathcal{T} \mid h = v \text{ or } t = v \}.
\end{equation}
The expanded subgraph is defined as:
\begin{equation}
    \mathcal{V}_{\text{expanded}} = \bigcup_{v \in \mathcal{S}} \{ h, t \mid (h, p, t) \in \mathcal{T}_{\text{one-hop}}(v) \},
\end{equation}
\begin{equation}
    \mathcal{P}_{\text{expanded}} = \bigcup_{v \in \mathcal{S}} \{ p \mid (h, p, t) \in \mathcal{T}_{\text{one-hop}}(v) \},
\end{equation}
\begin{equation}
    \mathcal{T}_{\text{expanded}} = \bigcup_{v \in \mathcal{S}} \mathcal{T}_{\text{one-hop}}(v).
\end{equation}
This forms the intermediate subgraph:
\begin{equation}
    \mathcal{G}_{\text{expanded}} = (\mathcal{V}_{\text{expanded}}, \mathcal{P}_{\text{expanded}}, \mathcal{T}_{\text{expanded}}),
\end{equation}
which may include off-topic entities and relationships.

\subsubsection{Steiner Tree Extraction}
The goal of extracting a \textit{Steiner tree} from $\mathcal{G}_{\text{expanded}}$ is to obtain a minimal yet structurally coherent subgraph that maintains essential connections between seed entities while discarding irrelevant nodes and edges. This step ensures that the final extracted subgraph is both informative and computationally manageable, preventing the inclusion of excessive noise that could degrade the quality of generated QA pairs.

We define the Steiner tree $\mathcal{T}_\mathcal{S}$ as the smallest connected subset of triples $\mathcal{T}$ from $\mathcal{T}_{\text{expanded}}$ that spans all seed entities $\mathcal{S}$. This is formulated as the following optimization problem:

\begin{equation}
    \mathcal{T}_\mathcal{S} = \arg\min_{\mathcal{T} \subseteq \mathcal{T}_{\text{expanded}}} \big( |\mathcal{T}| \big)
\end{equation}

subject to:

\begin{equation}
    \mathcal{S} \subseteq \mathcal{V}(\mathcal{T}),
\end{equation}

where $\mathcal{V}(\mathcal{T})$ represents the set of entities appearing in the selected triples $\mathcal{T}$.

To approximate the solution efficiently, we employ the algorithm proposed by \citet{mehlhorn1988faster}. However, since Steiner tree approximations do not always guarantee full connectivity among all seed entities -- especially if $\mathcal{G}_{\text{expanded}}$ itself is sparsely connected -- we introduce an additional filtering step to ensure that the extracted subgraph remains structurally coherent.

\subsubsection{Final Subgraph Selection}
While the Steiner tree extraction yields a compact subgraph, it may contain multiple disconnected components due to gaps in connectivity within $\mathcal{G}_{\text{expanded}}$. Since disconnected components hinder meaningful reasoning and QA generation, we refine the selection process by retaining only the largest connected component.

Given the connected components $\mathcal{C}_1, \mathcal{C}_2, \dots, \mathcal{C}k$ of $\mathcal{T}_\mathcal{S}$, the final subgraph is defined as:

\begin{equation} \mathcal{G}' = \arg\max_{\mathcal{C}i \subseteq \mathcal{T}\mathcal{S}} |\mathcal{V}(\mathcal{C}_i)| \end{equation}

where $\mathcal{V}(\mathcal{C}_i)$ denotes the set of entities in component $\mathcal{C}_i$.

Thus, the final extracted subgraph is:

\begin{equation} \mathcal{G}' = (\mathcal{V}', \mathcal{P}', \mathcal{T}'), \end{equation}

where: 
\begin{itemize} 
    \item $\mathcal{V}' = \mathcal{V}(\mathcal{G}')$ is the set of entities in the largest connected subgraph, 
    \item $\mathcal{P}' = \mathcal{P}(\mathcal{G}')$ is the set of predicates, 
    \item $\mathcal{T}'$ corresponds to the set of triples in $\mathcal{G}'$.
\end{itemize}

This selection ensures that $\mathcal{G}'$ maintains connectivity, preserving essential relationships for effective downstream applications such as QA generation.

\subsection{Generating QA Pairs}
\label{subsec:generatingQA}

Given a seed subgraph $\mathcal{G}'$, an LLM is used to generate QA pairs based on the information in the graph. First, $\mathcal{G}'$ is serialized as a list of triples in text form:

\[
    \mathcal{G}' = \{ (h, p, t) \mid h, t \in \mathcal{V}', \, p \in \mathcal{P}' \}.
\]

This serialized representation is then provided to the LLM using the prompt P1 in \ref{app:prompts}, instructing the model to generate questions based on facts in $\mathcal{G}'$, provide corresponding answers, and include the answer path (a subset of triples $\pi_i$) used to derive each answer.
QA pairs are generated through the function:

\begin{equation}
    \text{GenerateQA}(\mathcal{G}') \rightarrow \{(q_i, a_i, \pi_i)\}_{i=1}^N,
\end{equation}

where:
\begin{itemize}
    \item $q_i$: The $i$-th question,
    \item $a_i$: The corresponding answer,
    \item $\pi_i \subseteq \mathcal{T}'$: The answer path for the QA pair.
\end{itemize}

\begin{figure*}[h]
    \centering
    \begin{subfigure}{0.32\textwidth}
        \centering
        \includegraphics[width=\linewidth]{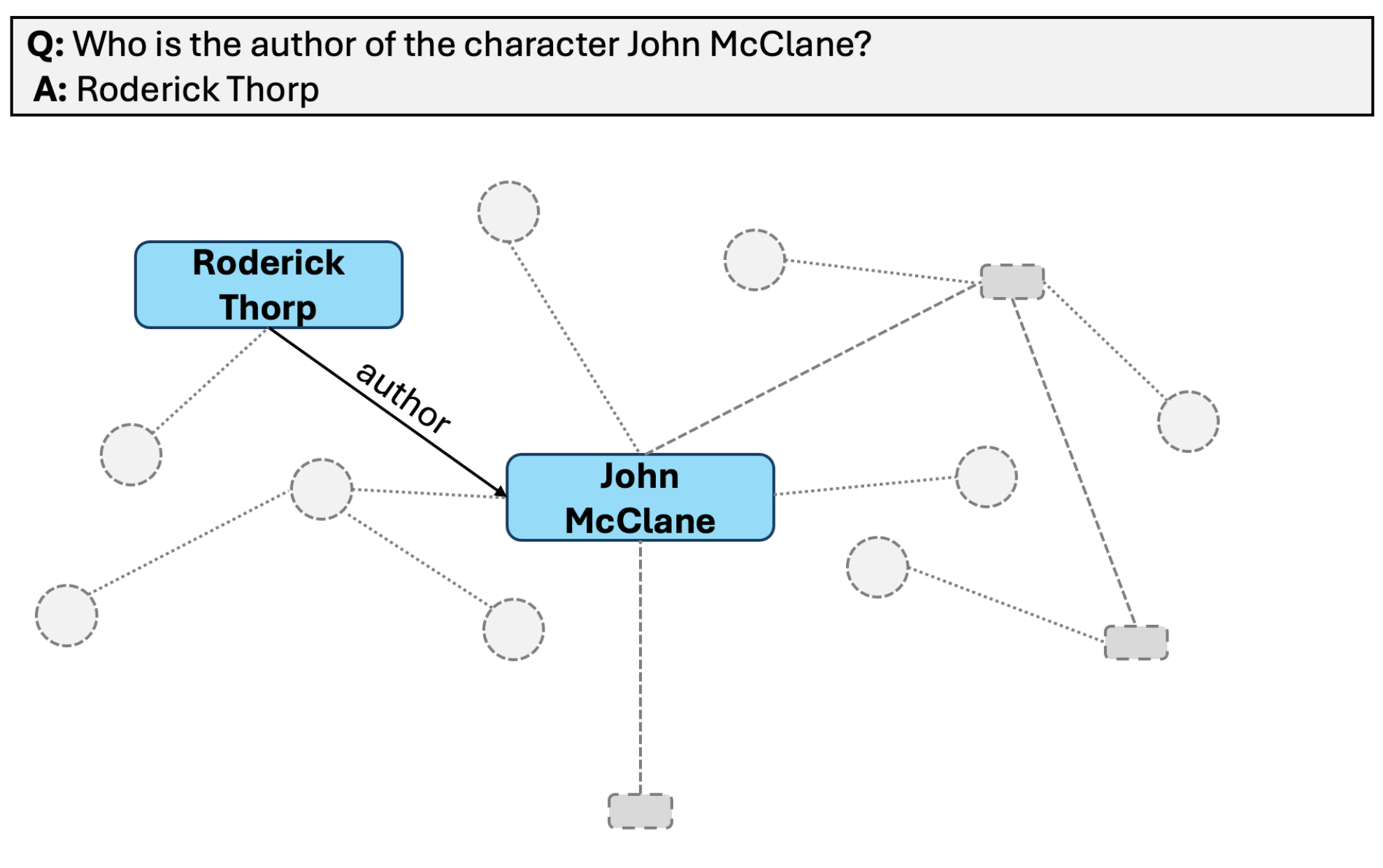}
        \caption{Answer Path 1}
    \end{subfigure}
    \hfill
    \begin{subfigure}{0.32\textwidth}
        \centering
        \includegraphics[width=\linewidth]{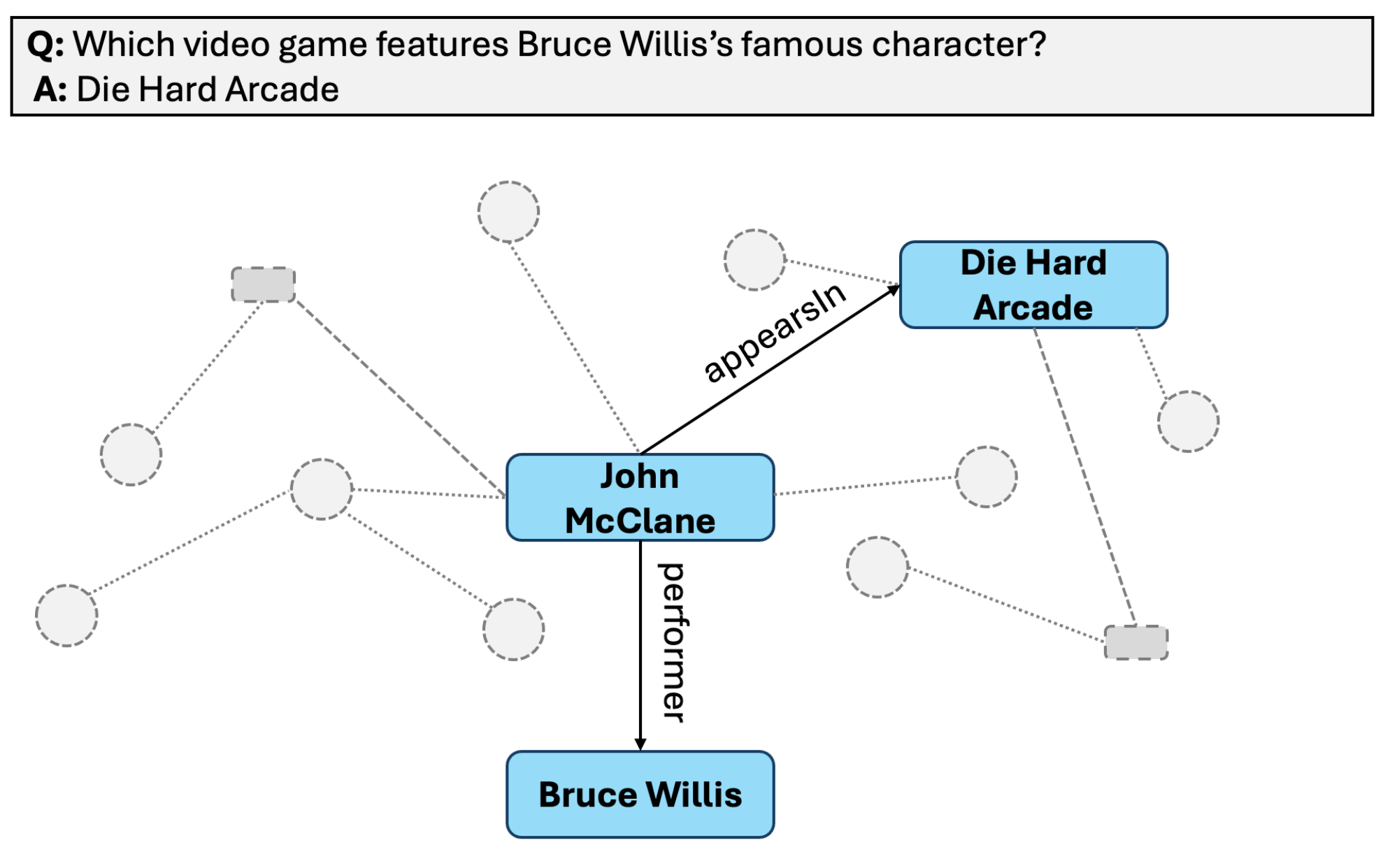}
        \caption{Answer Path 2}
    \end{subfigure}
    \hfill
    \begin{subfigure}{0.32\textwidth}
        \centering
        \includegraphics[width=\linewidth]{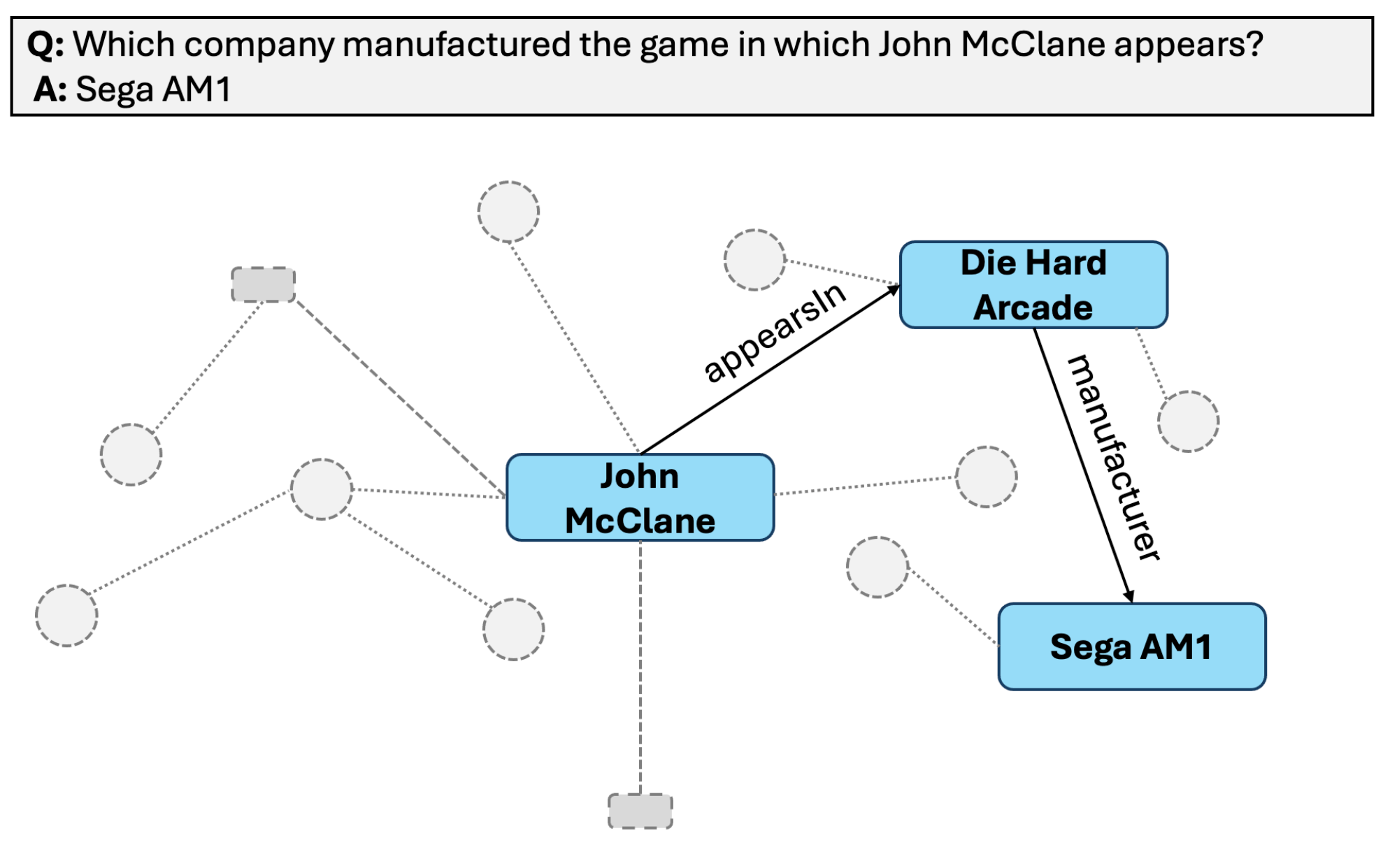}
        \caption{Answer Path 3}
    \end{subfigure}
    \caption{Illustration of how a single subgraph can generate multiple QA pairs. Each subfigure represents a different reasoning path within the same knowledge structure to answer diverse but related questions.}
    \label{fig:subgraph_variation}
\end{figure*}

A single subgraph can be leveraged to generate multiple QA pairs while remaining focused on the same topic. This is due to the interconnected nature of KGs, where entities and their relationships define different paths for retrieving relevant information. In Figure \ref{fig:subgraph_variation}, the subgraph centered around John McClane enables diverse but related questions, with variations in how entities are connected and how information needs to be retrieved.

For instance, the first question, ``Who is the author of John McClane?'', involves retrieving a directly connected entity through a single incoming edge. In contrast, the second question, ``Which video game features Bruce Willis’ famous character?'', requires reasoning over two outgoing edges independently, where different relationships link the central entity to multiple relevant nodes. The third question, ``Which company manufactured the game in which John McClane appears?'', requires multi-hop reasoning, as it involves first identifying an intermediate entity before retrieving the final answer.
These variations illustrate how the structure and directionality of edges influence the complexity of reasoning. Some questions rely on a single direct connection, while others necessitate broader exploration of multiple outgoing edges or sequential traversal across multiple nodes. Such structural differences impact retrieval strategies and highlight the challenges associated with knowledge-based question answering.

\subsection{Randomized QA Generation}
\label{subsec:randomizingQA}

To promote variability and make QA generation dynamic, we incorporate two parameters -- temperature ($T$) and triple reordering seed ($R$). The temperature parameter controls the degree of randomness in the LLM’s responses, with higher values promoting diverse outputs and lower values yielding more deterministic responses. Studies have shown that variations in input ordering can significantly impact generation outcomes \cite{Zhu2024DyVal2D,zong2023fool}. Leveraging this, we randomize the order of serialized triples in $\mathcal{G}'$ using a seed $R$, allowing reordering to serve as an additional mechanism for generating diverse outputs. To operationalize this, we extend \text{GenerateQA} by incorporating triple reordering and temperature control:

\begin{equation}
    \text{DynamicQA}(\mathcal{G}', T, R) = \text{GenerateQA}(\text{Reorder}(\mathcal{G}', R), T).
\end{equation}

Here, the function first applies $\text{Reorder}(\mathcal{G}', R)$, which randomizes the order of triples in $\mathcal{G}'$ using the seed $R$. The reordered triples are then passed to $\text{GenerateQA}$, where the LLM generates QA pairs along with answer paths with temperature $T$. 

By adjusting these parameters, the framework generates diverse QA samples from the same subgraph, ensuring variation while preserving meaningful relationships.

\subsection{Verifying QA Pairs}
\label{subsec:verifyQA}

To ensure that generated QA pairs are well-grounded in the subgraph $\mathcal{G}'$, a verification step is applied. This process checks whether the supporting path $\pi_{ij}$ of each QA pair $(q_{ij}, a_{ij}, \pi_{ij})$ exists within the extracted triples of $\mathcal{G}'$. The verification function is defined as:

\begin{equation}
    \text{VerifyPath}(\pi_{ij}, \mathcal{G}') =
    \begin{cases} 
    \text{True}, & \text{if } \pi_{ij} \subseteq \mathcal{T}' \\
    \text{False}, & \text{otherwise}.
    \end{cases}
\end{equation}

A QA pair is considered valid if all triples in $\pi_{ij}$ belong to $\mathcal{T}'$ and every referenced entity exists in $\mathcal{V}'$. Since LLMs can occasionally introduce errors or unsupported claims, performing these membership checks helps filter out hallucinated responses, ensuring that only factually grounded answers are retained.

\subsection{LLM-as-a-Judge for Annotations}
\label{subsec:annotationLLM}

To ensure that only high-quality QA pairs are retained in the final set, multiple LLMs are employed as automated annotators, reducing reliance on manual labeling. This enables real-time generation and evaluation of QA pairs, which is essential for dynamic assessment. Each QA pair is evaluated using two prompts: P2 for question assessment and P3 for answer verification (Please refer \S\ref{app:prompts} for full prompts).

Prompt P2 evaluates grammatical correctness and syntactic structure while detecting redundancy. It produces two binary flags: (1) Logical Structure, which is true if the question is well-formed, and (2) Redundancy, which is true if the question contains its own answer or is trivial. Prompt P3 assesses answer validity based on factual support and completeness, generating two additional flags: (3) Answer Support, which is true if supporting KG triples substantiate the answer, and (4) Answer Adequacy, which is true if the response fully addresses the question. Each flag is accompanied by a rationale, as prior work suggests that generating justifications improves performance \cite{wei2022chain,kojima2022large}, though only the final flag values are retained for filtering.

A QA pair is accepted if, across all three LLMs, it meets the following criteria: Logical Structure is true, Redundancy is false, Answer Support is true, and Answer Adequacy is true. To enforce a higher quality threshold, all LLM-as-a-Judge models must reach unanimous agreement on these criteria; any disagreement results in the QA pair being discarded.

\subsection{Dataset Construction and Artifacts}

The overall process for constructing \textsc{Dynamic-KGQA} is illustrated in Figure \ref{fig:pipeline}. We source seed texts from Wiki-40B \cite{guo2020wiki}, which aligns well with our requirements. Each entry in Wiki-40B corresponds to a Wikidata KG node and provides concise yet informative text focused on the respective node. Since we employ YAGO 4.5 \cite{suchanek2024yago}, which itself is derived from Wikidata, this choice ensures compatibility. Additionally, Wiki-40B offers large train, validation, and test splits, allowing us to create corresponding non-overlapping splits for \textsc{Dynamic-KGQA}, preventing any data leakage issues.

LLM selection is a design choice in dataset construction, as any suitable model can be used for QA generation and evaluation. In our setup, we use Claude 3.5 Sonnet v2 \cite{anthropic2024claude3} for the generation steps described in \S \ref{subsec:generatingQA} and \S \ref{subsec:randomizingQA}, as it is one of the state-of-the-art (SOTA) models at the time of writing. Following prior work \cite{schoenegger2024wisdom}, we construct a diverse LLM-as-a-Judge panel with models from different research labs. Specifically, we use Cohere's Command-R \cite{cohere2024commandr}, Mistral Small (24.02) \cite{mistral2025small3}, and Amazon Nova Lite \cite{Intelligence2024}.

For all models across all dataset construction steps, we use default parameters with a temperature of zero. Prompts are provided in \S \ref{appendix}.

\section{Dataset Characteristics}

\begin{table}[t]
    \centering
    \caption{Comparison of KGQA datasets. Size (n) denotes QA pairs per split. Base KG is the underlying knowledge base. QA Subgraph indicates subgraph availability. Dynamic shows support for generating consistent dynamic variants. \label{table:datasetComparison}}
    \resizebox{\columnwidth}{!}{%
    \begin{tabular}{lcccccc}
        \toprule
        \multirow{2}{*}{\textbf{Dataset}} & \multicolumn{3}{c}{\textbf{Size (n)}} & \multirow{2}{*}{\textbf{Base KG}} & \multirow{2}{*}{\makecell{\textbf{QA} \\ \textbf{Subgraph}}} & \multirow{2}{*}{\textbf{Dynamic}} \\
        \cmidrule(lr){2-4}
        & \textbf{Train} & \textbf{Test} & \textbf{Val} &  &  &  \\
        \toprule
        CWQ & 27,734 & 3,531 & - & Freebase & \makecell{\(\times\)} & \(\times\) \\
        WebQSP & 3,098 & 1,639 & - & Freebase & \makecell{\(\times\)} & \(\times\) \\
        GrailQA & 44,337 & 6,763 & 13,231 & Freebase & \makecell{\(\times\)} & \(\times\) \\
        QALD10-en & - & - & 394 & Wikidata & \makecell{\(\times\)} & \(\times\) \\
        Simple Questions & 75,910 & 21,687 & 10,845 & Freebase & \makecell{\(\times\)} & \(\times\) \\
        WebQuestions & 3,000 & 2,032 & 778 & Freebase & \makecell{\(\times\)} & \(\times\) \\
        \textsc{Dynamic-KGQA} & \textbf{200,000} & \textbf{40,000} & \textbf{40,000} & YAGO 4.5 & \makecell{\checkmark} & \checkmark \\
        \bottomrule
    \end{tabular}%
    }
\end{table}

\begin{table}[t]
    \centering
    \caption{Attributes of the \textsc{Dynamic-KGQA} dataset. \label{table:attributes}}
    \resizebox{\columnwidth}{!}{%
    \begin{tabular}{ll}
        \toprule
        \textbf{Name} & \textbf{Definition} \\
        \midrule
        \texttt{id} & Unique identifier for the QA pair. \\
        \texttt{question} & Input question text. \\
        \texttt{answer} & YAGO entity name. \\
        \texttt{answer\_readable} & Human-readable version of the answer. \\
        \texttt{answer\_uri} & YAGO entity URI. \\
        \texttt{supporting\_facts} & Triples supporting the answer. \\
        \texttt{supporting\_facts\_uri} & YAGO URI for supporting facts. \\
        \texttt{subgraph} & Subgraph for the QA pair. \\
        \texttt{subgraph\_size} & Size of the subgraph. \\
        \texttt{logical\_structure\_flag\_n} & Flag for logical structure (LLM-as-a-Judge \(n\)). \\
        \texttt{logical\_structure\_reasoning\_n} & Explanation for the logical structure flag \(n\). \\
        \texttt{redundancy\_flag\_n} & Flag for redundant info (LLM-as-a-Judge \(n\)). \\
        \texttt{redundancy\_reasoning\_n} & Explanation for the redundancy flag \(n\). \\
        \texttt{answer\_support\_flag\_n} & Flag for answer support (LLM-as-a-Judge \(n\)). \\
        \texttt{answer\_support\_reasoning\_n} & Explanation for the support flag \(n\). \\
        \texttt{answer\_adequacy\_flag\_n} & Flag for answer adequacy (LLM-as-a-Judge \(n\)). \\
        \texttt{answer\_adequacy\_reasoning\_n} & Explanation for the adequacy flag \(n\). \\
        \bottomrule
    \end{tabular}%
    }
\end{table}

In this section, we outline the key attributes of the \textsc{Dynamic-KGQA} dataset and compare its characteristics with other commonly used KGQA datasets. Table \ref{table:attributes} presents the attributes included for each datapoint in the dataset.

\vspace{2mm}
\noindent{\textbf{\textit{Compatibility for Automated Reasoning.}}}
While recent KGQA methods leverage the latest language models with advanced reasoning capabilities, many KGQA benchmarks still rely on Freebase \cite{bollacker2008freebase}, a knowledge base that was discontinued in 2015 and is no longer maintained or updated \cite{pellissier2016freebase}. As a result, these methods, despite being SOTA, are evaluated on data that is more than a decade old. Wikidata is a more up-to-date alternative as it is a collaboratively maintained knowledge base. However, Wikidata introduces challenges such as taxonomic loops, redundant relations, and contradictions, which can hinder automated reasoning.

To address these limitations, Yago 4.5 \cite{suchanek2024yago} provides a more structured alternative by eliminating taxonomic inconsistencies, reducing redundancy, and ensuring a contradiction-free ontology. These improvements make Yago 4.5 particularly well-suited for automated reasoning and KGQA tasks that depend on structured knowledge. Based on these considerations, we build \textsc{Dynamic-KGQA} on Yago 4.5, making it more suitable for modern KGQA methods that rely on logical inference and knowledge exploration.

\vspace{2mm}
\noindent{\textbf{\textit{Localized Subgraph Support.}}}
Several methods rely on the strategy of extracting subgraphs, as they constrain the search space to entities and relations directly pertinent to the query \cite{zhang2022subgraph,ding2024enhancing,jiang2022unikgqa,sun2019pullnet,sun2018open,saxena2020improving,jiang2023reasoninglm,wu2023retrieve,zhang2018variational}. Traversing the entire KG introduces exponential complexity growth as graph diameter increases, which forces models to process numerous irrelevant edges, thereby obscuring critical reasoning paths. However, since existing KGQA datasets do not provide corresponding subgraphs for each QA pair, it becomes infeasible to study whether the bottleneck of KGQA methods lies in subgraph extraction or other parts of the method. \citet{ding2024enhancing} discuss the shortcomings of current subgraph extraction methods, which underestimate the importance of structural dependencies among evidence facts. From an evaluation perspective, localized subgraphs enable controlled testing environments that isolate specific reasoning capabilities. \textsc{Dynamic-KGQA} provides a localized subgraph for each QA pair, significantly enhancing both the development and evaluation of KGQA methods.

\vspace{2mm}
\noindent{\textbf{\textit{Scale.}}}
As shown in Table~\ref{table:datasetComparison}, \textsc{Dynamic-KGQA} is significantly larger than most existing datasets used for evaluating KGQA methods. The combination of its size and diverse topic distribution enables a comprehensive benchmark for question answering over structured data. Prior research indicates that scaling training data leads to consistent improvements in performance \cite{hoffmann2022training,chen2024scaling}. While architectural and methodological advancements contribute to performance gains, increasing training data remains a fundamental factor in enhancing model generalization and efficiency.

Additionally, certain existing datasets such as CWQ \cite{talmor2018web} and WebQSP \cite{yih2016value} contain only train and test splits, lacking a dedicated validation set. The importance of a three-way dataset split -- training, validation, and test -- has become increasingly evident as machine learning models grow more complex. Introducing a dedicated validation split helps mitigate the risk of overfitting to the test set \cite{Goodfellow-et-al-2016,bishop2006pattern}. \citet{bishop2006pattern} highlight the role of validation split in hyperparameter tuning and model selection, ensuring that the test set remains reserved for final benchmarking. Informed by these principles, \textsc{Dynamic-KGQA} includes sufficiently large and well-separated splits for all three phases of model development.

\vspace{2mm}
\noindent{\textbf{\textit{Dynamic Samples.}}}
All three splits of \textsc{Dynamic-KGQA} support dynamic samples, as each QA pair in each split has a corresponding subgraph that facilitates the generation of QA pairs within the same topic. Variants of the training split can be used for data augmentation in the training phase, a well-established practice that helps alleviate robustness issues, spurious correlations, and other concerns beyond enhancing performance \cite{morris2020textattack,goel2020model,goodfellow2014explaining}. Furthermore, \citet{Zhu2023DyValDE} show that dynamic samples can improve the abilities of LLMs for downstream tasks. Using dynamic sampling for the validation split may also address concerns such as overfitting. For benchmarking purposes, especially in the era of LLMs where any data available on the web gets consumed by these models, dynamic samples provide an effective solution to issues such as data contamination and memorization \cite{Zhu2023DyValDE,Zhu2024DyVal2D}.

\begin{table*}[h]
    \centering

    \caption{LLM baselines on multiple QA datasets, including \textsc{Dynamic-KGQA}. Models are evaluated using standard prompting (IO), Chain-of-Thought prompting (CoT), and Think-on-Graph (ToG), establishing baselines across datasets. \label{table:scores}}
    
    \resizebox{\textwidth}{!}{%
    \begin{tabular}{lccccccc}
        \toprule
        Method & \multicolumn{4}{c}{Multi-Hop KBQA} & \multicolumn{1}{c}{Single-Hop KBQA} & \multicolumn{1}{c}{Open-Domain QA} & \multicolumn{1}{c}{Multihop Dynamic KGQA} \\
        \cmidrule(lr){2-5} \cmidrule(lr){6-6} \cmidrule(lr){7-7} \cmidrule(lr){8-8}
        & CWQ & WebQSP & GrailQA & QALD10-en & Simple Questions & WebQuestions & \textsc{Dynamic-KGQA} \\
        \midrule
        Llama 3.2 3B Ins - IO & 17.98 & 31.97 & 13.1 & 36.34 & 12.4 & 27.12 & 12.03 \\
        Llama 3.2 3B Ins - CoT & 25.43 & 46.92 & 21.2 & 44.44 & 16.8 & 41.39 & 16.07 \\
        Command-R - IO & 24.36 & 48.38 & 19.8 & 36.04 & 18.9 & 41.73 & 16.88 \\
        Command-R - CoT & 28.72 & 58.99 & 28.2 & 42.94 & 22.6 & 51.97 & 18.45 \\
        Mistral Small (24.02) - IO & 29.09 & 51.86 & 25.6 & 38.74 & 19.7 & 44.59 & 19.49 \\
        Mistral Small (24.02) - CoT & 37.10 & 60.65 & 32.7 & 46.85 & 24.1 & 53.44 & 22.18 \\
        Claude 3.5 Sonnet v2 - IO & 30.16 & 51.68 & 26.0 & 36.64 & 25.3 & 44.78 & 19.78 \\
        Claude 3.5 Sonnet v2 - CoT & 40.02 & 65.53 & 36.7 & 52.55 & 33.5 & 57.23 & 25.22 \\
        GPT-4o-mini - IO & 28.18 & 49.36 & 25.9 & 36.64 & 22.2 & 43.26 & 18.50 \\
        GPT-4o-mini - CoT & 35.06 & 61.62 & 32.9 & 44.14 & 25.8 & 54.13 & 21.14 \\
        GPT-4o - IO & 33.81 & 53.57 & 27.0 & 39.64 & 26.0 & 45.57 & 24.44 \\
        GPT-4o - CoT & 42.11 & 62.17 & 36.3 & 46.85 & 31.7 & \textbf{54.53} & 27.43 \\
        GPT-4o - ToG & \textbf{44.22} & \textbf{67.42} & \textbf{39.8} & \textbf{54.65} & \textbf{34.4} & 53.67 & \textbf{27.80} \\
        \bottomrule
    \end{tabular}%
    }
\end{table*}

\begin{table}[h]
    \centering
    \caption{Pairwise Q\&A matching across dynamic samples (Runs 1–3). $\boldsymbol{\chi^2}$ and p-values assess topic label distributions via Chi-Square tests. Cramer's V ($\boldsymbol{\phi_c}$) quantifies effect size. \label{table:dynamic}}
    \resizebox{\columnwidth}{!}{%
    \begin{tabular}{lcccccc}
        \toprule
        \textbf{Comparison} & \textbf{Identical} & \textbf{Paraphrased} & \textbf{Unique} & $\boldsymbol{\chi^2}$ & \textbf{p-value} & \textbf{$\boldsymbol{\phi_c}$} \\
        \midrule
        Run 1 vs 2 & 8   & 308   & 1684  & 3.33  & 0.6489 & 0.0288 \\
        Run 1 vs 3 & 12  & 320   & 1668  & 1.86  & 0.8679 & 0.0216 \\
        Run 2 vs 3 & 15  & 313   & 1672  & 3.45  & 0.6311 & 0.0293 \\
        \bottomrule
    \end{tabular}%
    }
\end{table}


\section{Experiments and Results}

In this section, we present the results of our experiments and establish initial baselines for \textsc{Dynamic-KGQA}, providing a reference point for future research. To ensure diversity, we evaluate recent LLMs from multiple research labs. Additionally, we conduct statistical tests to evaluate the consistency of dynamic samples across runs, examining whether they belong to the same topic distribution.

We report baseline scores for various LLMs, including Llama 3.2 3B \cite{dubey2024llama}, Cohere's Command-R \cite{cohere2024commandr}, Mistral Small \cite{mistral2025small3}, Claude 3.5 Sonnet v2 \cite{anthropic2024claude3}, GPT-4o-mini \cite{hurst2024gpt}, and GPT-4o \cite{hurst2024gpt}. Each model is evaluated using standard prompting (IO prompt) \cite{brown2020language} and Chain-of-Thought prompting (CoT prompt) \cite{wei2022chain}. Additionally, we report the performance of Think-on-Graph (ToG) \cite{sun2024thinkongraph}, a KGQA method that leverages an LLM agent and has previously reported SOTA results on multiple KGQA benchmarks. 

For comparative analysis, we include widely used KGQA datasets in our experiments, such as CWQ \cite{talmor2018web}, WebQSP \cite{yih2016value}, GrailQA \cite{gu2021beyond}, QALD10-en \cite{usbeck2024qald}, Simple Questions \cite{bordes2015large}, Web Questions \cite{berant2013semantic}, and \textsc{Dynamic-KGQA}. For all the previously established datasets, we use the same splits as in \cite{sun2024thinkongraph}. A detailed comparison of these datasets is provided in Table \ref{table:scores}.

All rows in Table \ref{table:scores} with IO and CoT prompting techniques reflect scores based on the LLM's inherent knowledge, without relying on external sources. Notably, for most datasets, our reported scores for OpenAI models are higher than or comparable to those in \cite{sun2024thinkongraph}, despite using the same prompting strategies. This increase in scores can be attributed to model improvements, as \citet{sun2024thinkongraph} used GPT-3.5-turbo and GPT-4, whereas our experiments leverage the more recent GPT-4o and GPT-4o-mini. Another notable result is that Claude 3.5 Sonnet v2 performed competitively with GPT-4o, sometimes matching or even exceeding its performance. In contrast, the smallest model in our experiments, Llama 3.2 3B, consistently reported the lowest scores across all datasets, suggesting a strong correlation between model size and performance on knowledge-intensive tasks.

As expected, GPT-4o-ToG consistently achieves the highest scores across most datasets, as it incorporates structured knowledge from a KG, whereas other models rely solely on their internal knowledge (see Table \ref{table:scores}). While these improvements are significant, the gap between ToG and LLM-only performance is smaller than reported in \cite{sun2024thinkongraph}. This discrepancy could be due to several factors. First, we use GPT-4o instead of GPT-3.5-turbo or GPT-4, as OpenAI is phasing out these older models through increased pricing, making them less feasible due to resource constraints. Prior studies indicate that even with the same LLM, performance on specific tasks can degrade over time due to periodic updates \cite{chen2023chatgpt}. Second, ToG is specifically designed for Freebase \cite{bollacker2008freebase}, using SPARQL queries, algorithmic refinements, and prompt optimizations tailored to that KG. In contrast, our experiments are conducted using YAGO 4.5 \cite{suchanek2024yago}, which differs significantly in schema and structure. These differences may contribute to the observed performance variations. We leave further exploration of knowledge integration techniques, prompt optimizations, and model tuning for future work.

As seen in Table \ref{table:scores}, scores on \textsc{Dynamic-KGQA} are consistently lower compared to older datasets. A likely reason for this is that, at the time of writing, \textsc{Dynamic-KGQA} has not been incorporated into any LLM’s training data, meaning it remains unaffected by data contamination and memorization effects. Moreover, since we provide a mechanism to generate new dynamic QA samples as variants of the same test split, \textsc{Dynamic-KGQA} can continue to remain immune from contamination over time. This provides an opportunity for new KGQA methods to demonstrate improvements, where performance gains would more directly reflect reasoning and generalization rather than retrieval from pre-trained knowledge. Additionally, the dataset’s challenging nature suggests that it could serve as a valuable benchmark for evaluating KGQA models under more realistic, unseen conditions.

\begin{figure*}[h]
    \centering
    \begin{subfigure}{0.19\textwidth}
        \centering
        \includegraphics[width=\linewidth]{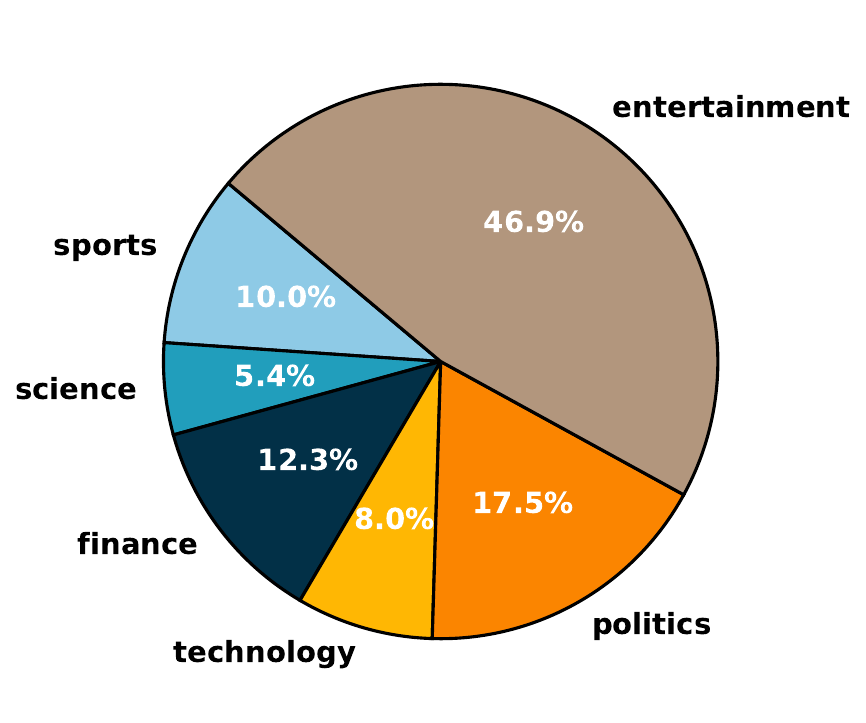}
        \caption{CWQ}
    \end{subfigure}
    \hfill
    \begin{subfigure}{0.19\textwidth}
        \centering
        \includegraphics[width=\linewidth]{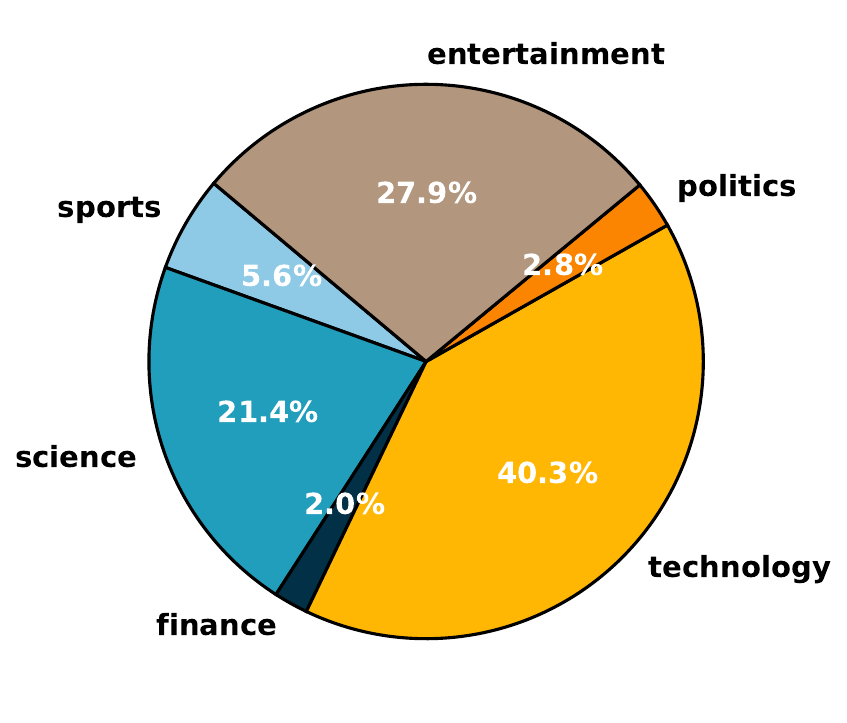}
        \caption{GrailQA}
    \end{subfigure}
    \hfill
    \begin{subfigure}{0.19\textwidth}
        \centering
        \includegraphics[width=\linewidth]{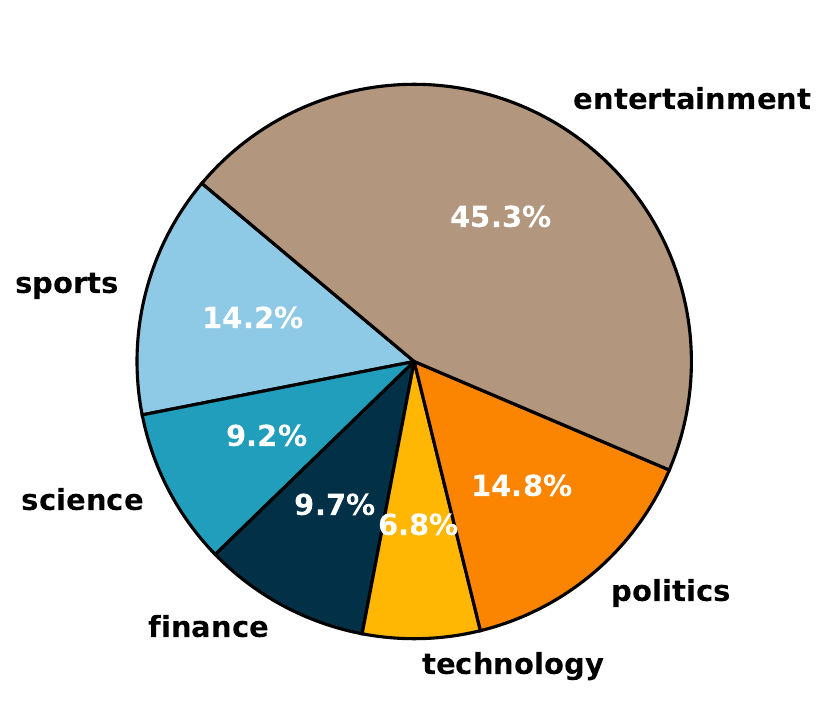}
        \caption{WebQSP}
    \end{subfigure}
    \hfill
    \begin{subfigure}{0.19\textwidth}
        \centering
        \includegraphics[width=\linewidth]{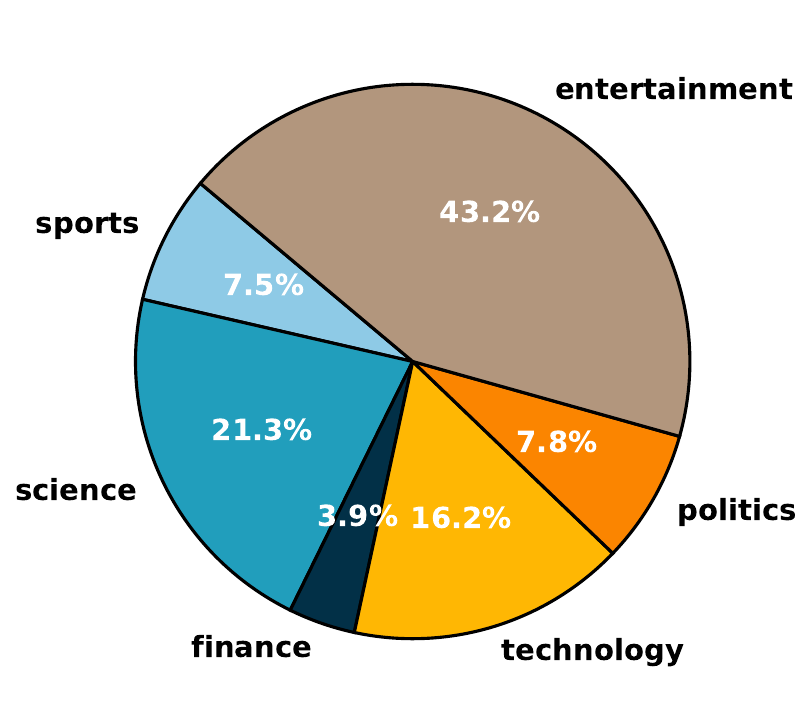}
        \caption{QALD10-en}
    \end{subfigure}
    \hfill
    \begin{subfigure}{0.19\textwidth}
        \centering
        \includegraphics[width=\linewidth]{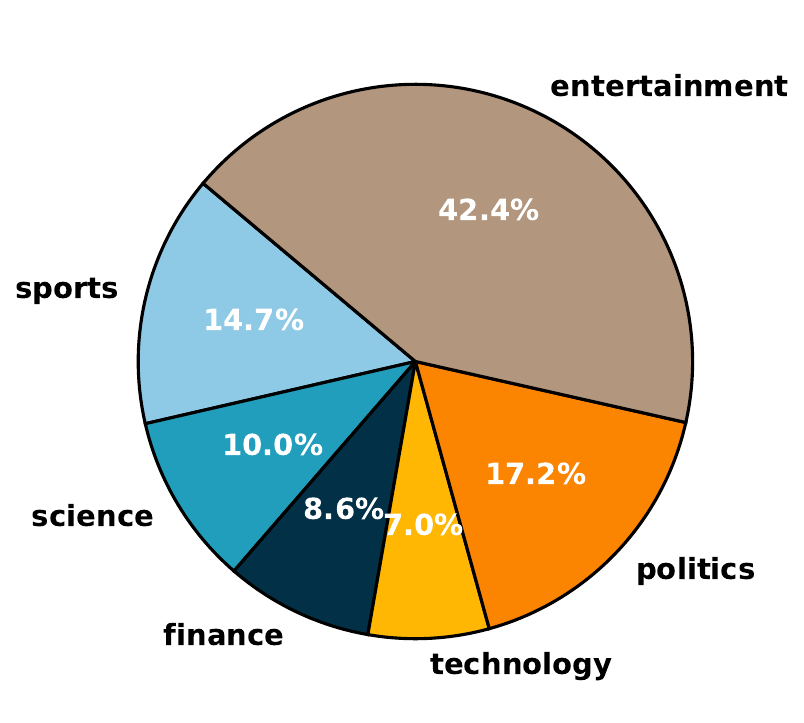}
        \caption{WebQuestions}
    \end{subfigure}

    \vspace{1em}

    \begin{subfigure}{0.19\textwidth}
        \centering
        \includegraphics[width=\linewidth]{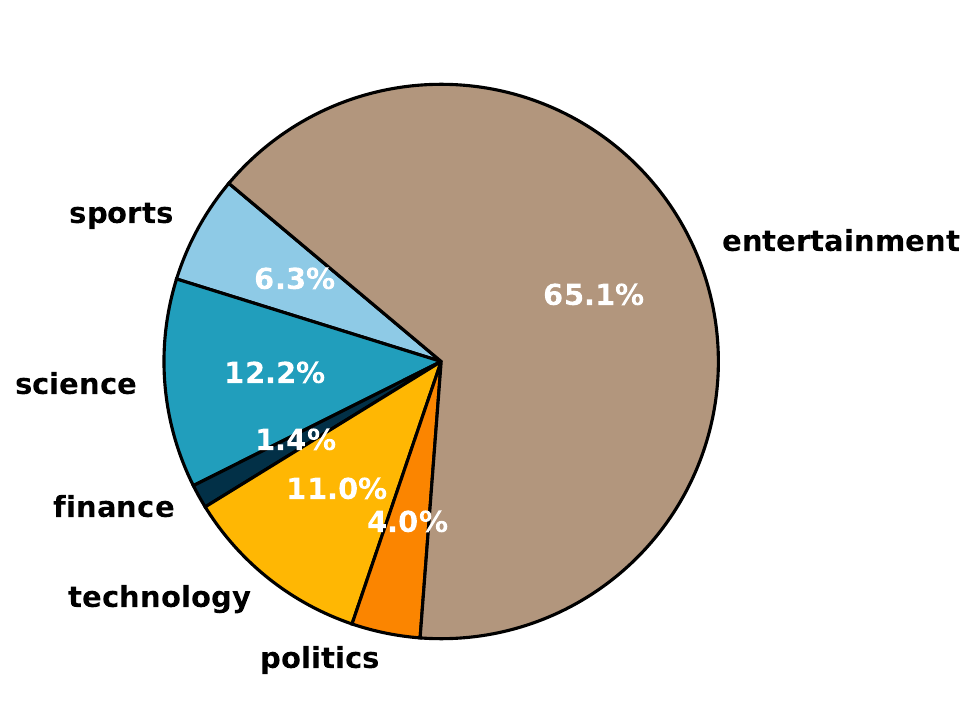}
        \caption{Simple Questions}
    \end{subfigure}
    \hfill
    \begin{subfigure}{0.19\textwidth}
        \centering
        \includegraphics[width=\linewidth]{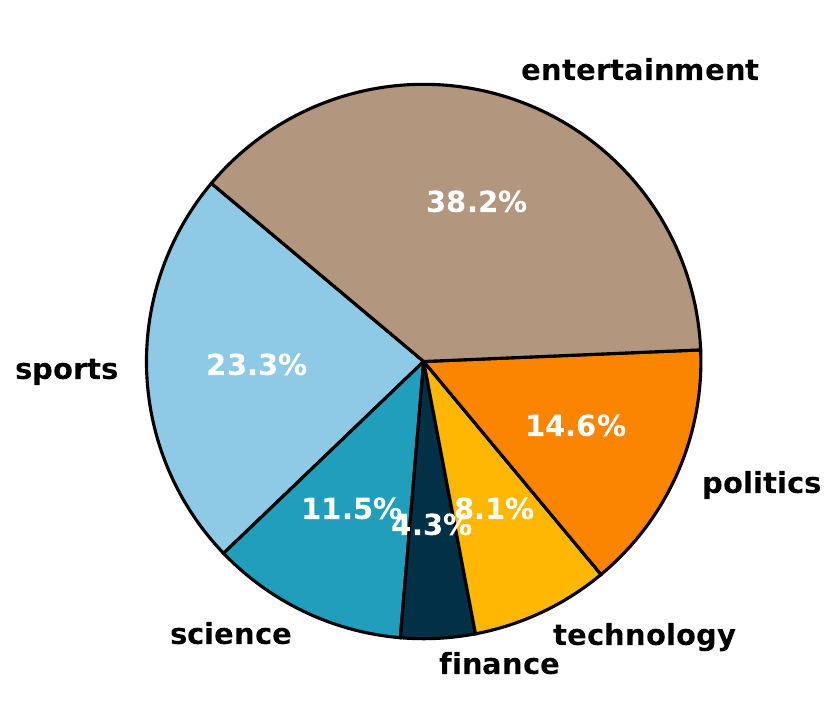}
        \caption{DynamicKGQA}
    \end{subfigure}
    \hfill
    \begin{subfigure}{0.19\textwidth}
        \centering
        \includegraphics[width=\linewidth]{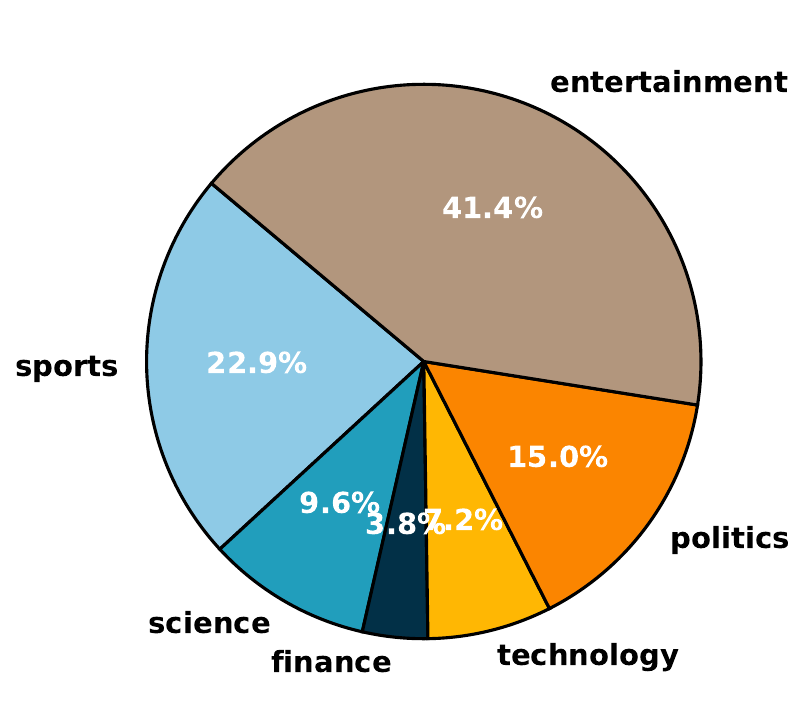}
        \caption{DynamicKGQA 2K-R1}
    \end{subfigure}
    \hfill
    \begin{subfigure}{0.19\textwidth}
        \centering
        \includegraphics[width=\linewidth]{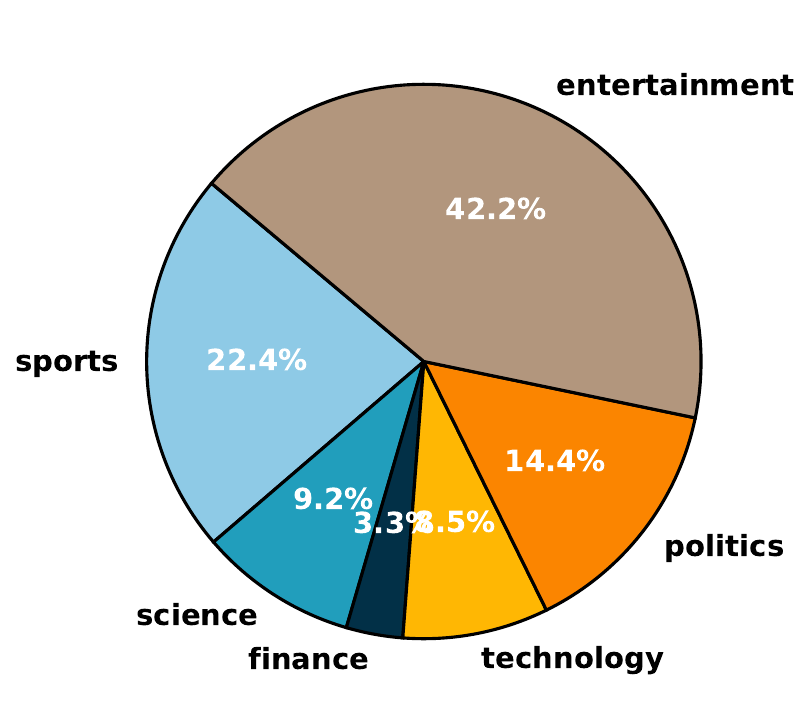}
        \caption{DynamicKGQA 2K-R2}
    \end{subfigure}
    \hfill
    \begin{subfigure}{0.19\textwidth}
        \centering
        \includegraphics[width=\linewidth]{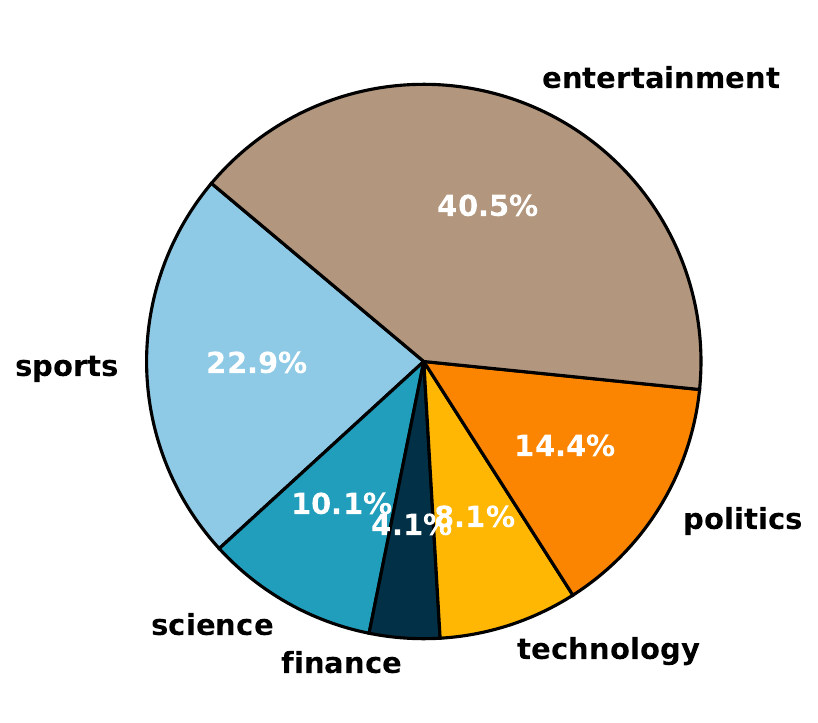}
        \caption{DynamicKGQA 2K-R3}
    \end{subfigure}

    \caption{Topic distribution of KGQA datasets.}
    \label{fig:topic_distribution}
    \vspace{-2mm}
\end{figure*}

To evaluate the consistency of dynamic samples across runs, we randomly sample 2,000 datapoints from the 40,000 test split. This smaller split helps keep the experiments within budget constraints. For each of these 2,000 datapoints, we generate three distinct dynamic sample variants, each using a unique triple reordering seed ($R$) per run, with the temperature ($T$) set to 0.8. As shown in Table \ref{table:dynamic}, this process results in three variants, with fewer than 0.01\% of QA pairs being identical, approximately 16\% paraphrased QA pairs, and the remaining instances consisting of unique QA pairs on the same topic. These findings indicate sufficient variability across runs, reducing the risk of memorization.

Furthermore, to assess the consistency of topic distribution across different runs, we generated topic labels for each QA pair using BART \cite{lewis2019bart}, categorizing them into six broad domains: sports, science, finance, technology, politics, and entertainment. We then conducted a Chi-Square test to determine whether the topic distributions of different dynamic variants were statistically similar. Formally, we define the null hypothesis (\(H_0\)) as the assumption that topic distributions across runs are drawn from the same underlying distribution, indicating no significant variation. The alternative hypothesis (\(H_1\)) posits that the distributions differ significantly. To determine statistical significance, we set a significance threshold of \(\alpha=0.05\), meaning that a p-value below \(0.05\) would indicate sufficient evidence to reject \(H_0\)  and conclude that the distributions are significantly different. 

As shown in Table \ref{table:datasetComparison}, the \(\chi^2\) statistics remain low across all pairwise comparisons, with corresponding p-values exceeding conventional significance thresholds (\( p > 0.05 \)), indicating insufficient evidence to reject the null hypothesis. We thus report the Cramer's V effect size across all pairwise comparisons, to quantify the magnitude of potential deviations. Across all comparisons, the Cramer's V remains consistently low (\( \phi_c < 0.03 \)), which suggests negligible variation in the topic distributions across different runs. The distributional stability is further supported by the visualizations in Figure \ref{fig:topic_distribution} (h,i,j), where category proportions remain nearly invariant across generated samples. Additionally, in Figure \ref{fig:topic_distribution}, we observe that \textsc{Dynamic-KGQA} maintains a more balanced distribution across categories compared to a few older datasets. For instance, Simple Questions is heavily skewed toward entertainment (65.1\%), while GrailQA overrepresents technology. In contrast, \textsc{Dynamic-KGQA} ensures a more even spread across topics, making it a more generalizable benchmark for KGQA models.

\section{Limitations}

\vspace{2mm}
\noindent{\textbf{\textit{KG Coverage.}}}

Since our framework relies on a YAGO 4.5 \cite{suchanek2024yago}, any inherent limitations of the base KG may propagate. While KGs offer advantages such as structured representation, editability, and auditability, they remain constrained by factors such as periodic updates and potential gaps in coverage. However, this limitation is not unique to our approach but is common across all KGQA methods and datasets. In our work, we attempt to mitigate this concern by leveraging the latest KG best suited for automated reasoning.

\vspace{2mm}
\noindent{\textbf{\textit{LLM Limitations. }}}
We employ LLMs for both data generation and evaluation. While these models have demonstrated strong capabilities, they are also prone to errors, including hallucinations and factual inconsistencies \cite{xu2024hallucinationinevitableinnatelimitation,zhang2023siren,huang2023survey,Li2023HaluEvalAL}. However, LLMs have also shown effectiveness in tasks such as fact verification \cite{guan2023language,min2023factscore}, and some studies suggest they can even outperform human annotators in specific contexts \cite{gilardi2023chatgpt,ostyakova2023chatgpt}. In dynamic evaluation paradigms requiring on-demand dataset generation, the adoption of automated pipelines and LLMs is often necessary.

\vspace{2mm}
\noindent{\textbf{\textit{Baseline Evaluation Constraints. }}}
Although we report results across multiple models, prompting techniques, and a SOTA KGQA method, many alternative configurations remain unexplored. The continuous emergence of new methods necessitates further experimentation, including fine-tuning and hyperparameter optimization, to achieve optimal performance. However, conducting such extensive evaluations has significant computational and cost implications, limiting the feasibility of broader model selection and refinement.

\vspace{-2mm}

\section{Conclusion}

In this work, we introduced \textsc{Dynamic-KGQA}, a scalable and dynamic QA benchmarking framework. To the best of our knowledge, it is the first dynamic benchmark specifically designed for KGQA. Unlike static benchmarks, which are susceptible to data contamination and memorization, \textsc{Dynamic-KGQA} produces adaptive QA pairs while maintaining statistical consistency across runs, enabling a more rigorous and reliable evaluation of KGQA systems. Our framework addresses key limitations of existing benchmarks by providing compact, semantically coherent subgraphs for each QA pair, facilitating controlled testing environments and explicit development of KGQA capabilities. Overall, \textsc{Dynamic-KGQA} represents a meaningful step toward more adaptable and contamination-resistant QA benchmarking.


\section*{Acknowledgements}

We thank Carter Swartout for his valuable discussions and explorations during the initial phase of this project. This work was partially supported by cloud computing credits provided through an Amazon Research Award to Dr. Chirag Shah, titled ``Fairness as a Service: Operationalizing Fairness in Search and Recommendation Applications Through a Novel Multi-Objective Optimization Framework," and by the University of Washington eScience Institute in partnership with Microsoft Azure.

\balance
\bibliographystyle{ACM-Reference-Format}
\bibliography{references}


\begin{thebibliography}{80}


\ifx \showCODEN    \undefined \def \showCODEN     #1{\unskip}     \fi
\ifx \showDOI      \undefined \def \showDOI       #1{#1}\fi
\ifx \showISBNx    \undefined \def \showISBNx     #1{\unskip}     \fi
\ifx \showISBNxiii \undefined \def \showISBNxiii  #1{\unskip}     \fi
\ifx \showISSN     \undefined \def \showISSN      #1{\unskip}     \fi
\ifx \showLCCN     \undefined \def \showLCCN      #1{\unskip}     \fi
\ifx \shownote     \undefined \def \shownote      #1{#1}          \fi
\ifx \showarticletitle \undefined \def \showarticletitle #1{#1}   \fi
\ifx \showURL      \undefined \def \showURL       {\relax}        \fi
\providecommand\bibfield[2]{#2}
\providecommand\bibinfo[2]{#2}
\providecommand\natexlab[1]{#1}
\providecommand\showeprint[2][]{arXiv:#2}

\bibitem[Anthropic(2024)]%
        {anthropic2024claude3}
\bibfield{author}{\bibinfo{person}{Anthropic}.} \bibinfo{year}{2024}\natexlab{}.
\newblock \bibinfo{booktitle}{\emph{The Claude 3 Model Family: Opus, Sonnet, Haiku}}.
\newblock
\urldef\tempurl%
\url{https://www-cdn.anthropic.com/de8ba9b01c9ab7cbabf5c33b80b7bbc618857627/Model_Card_Claude_3.pdf}
\showURL{%
\tempurl}
\newblock
\shownote{Accessed: 2025-02-16}.


\bibitem[Bao et~al\mbox{.}(2016)]%
        {bao2016constraint}
\bibfield{author}{\bibinfo{person}{Junwei Bao}, \bibinfo{person}{Nan Duan}, \bibinfo{person}{Zhao Yan}, \bibinfo{person}{Ming Zhou}, {and} \bibinfo{person}{Tiejun Zhao}.} \bibinfo{year}{2016}\natexlab{}.
\newblock \showarticletitle{Constraint-based question answering with knowledge graph}. In \bibinfo{booktitle}{\emph{Proceedings of COLING 2016, the 26th international conference on computational linguistics: technical papers}}. \bibinfo{pages}{2503--2514}.
\newblock


\bibitem[Berant et~al\mbox{.}(2013)]%
        {berant2013semantic}
\bibfield{author}{\bibinfo{person}{Jonathan Berant}, \bibinfo{person}{Andrew Chou}, \bibinfo{person}{Roy Frostig}, {and} \bibinfo{person}{Percy Liang}.} \bibinfo{year}{2013}\natexlab{}.
\newblock \showarticletitle{Semantic parsing on freebase from question-answer pairs}. In \bibinfo{booktitle}{\emph{Proceedings of the 2013 conference on empirical methods in natural language processing}}. \bibinfo{pages}{1533--1544}.
\newblock


\bibitem[Berant and Liang(2014)]%
        {berant2014semantic}
\bibfield{author}{\bibinfo{person}{Jonathan Berant} {and} \bibinfo{person}{Percy Liang}.} \bibinfo{year}{2014}\natexlab{}.
\newblock \showarticletitle{Semantic parsing via paraphrasing}. In \bibinfo{booktitle}{\emph{Proceedings of the 52nd Annual Meeting of the Association for Computational Linguistics (Volume 1: Long Papers)}}. \bibinfo{pages}{1415--1425}.
\newblock


\bibitem[Bishop and Nasrabadi(2006)]%
        {bishop2006pattern}
\bibfield{author}{\bibinfo{person}{Christopher~M Bishop} {and} \bibinfo{person}{Nasser~M Nasrabadi}.} \bibinfo{year}{2006}\natexlab{}.
\newblock \bibinfo{booktitle}{\emph{Pattern recognition and machine learning}}. Vol.~\bibinfo{volume}{4}.
\newblock \bibinfo{publisher}{Springer}.
\newblock


\bibitem[Bollacker et~al\mbox{.}(2008)]%
        {bollacker2008freebase}
\bibfield{author}{\bibinfo{person}{Kurt Bollacker}, \bibinfo{person}{Colin Evans}, \bibinfo{person}{Praveen Paritosh}, \bibinfo{person}{Tim Sturge}, {and} \bibinfo{person}{Jamie Taylor}.} \bibinfo{year}{2008}\natexlab{}.
\newblock \showarticletitle{Freebase: a collaboratively created graph database for structuring human knowledge}. In \bibinfo{booktitle}{\emph{Proceedings of the 2008 ACM SIGMOD international conference on Management of data}}. \bibinfo{pages}{1247--1250}.
\newblock


\bibitem[Bordes et~al\mbox{.}(2015)]%
        {bordes2015large}
\bibfield{author}{\bibinfo{person}{Antoine Bordes}, \bibinfo{person}{Nicolas Usunier}, \bibinfo{person}{Sumit Chopra}, {and} \bibinfo{person}{Jason Weston}.} \bibinfo{year}{2015}\natexlab{}.
\newblock \showarticletitle{Large-scale simple question answering with memory networks}.
\newblock \bibinfo{journal}{\emph{arXiv preprint arXiv:1506.02075}} (\bibinfo{year}{2015}).
\newblock


\bibitem[Brown(2020)]%
        {brown2020language}
\bibfield{author}{\bibinfo{person}{Tom~B Brown}.} \bibinfo{year}{2020}\natexlab{}.
\newblock \showarticletitle{Language models are few-shot learners}.
\newblock \bibinfo{journal}{\emph{arXiv preprint arXiv:2005.14165}} (\bibinfo{year}{2020}).
\newblock


\bibitem[Chen et~al\mbox{.}(2024b)]%
        {chen2024planongraph}
\bibfield{author}{\bibinfo{person}{Liyi Chen}, \bibinfo{person}{Panrong Tong}, \bibinfo{person}{Zhongming Jin}, \bibinfo{person}{Ying Sun}, \bibinfo{person}{Jieping Ye}, {and} \bibinfo{person}{Hui Xiong}.} \bibinfo{year}{2024}\natexlab{b}.
\newblock \showarticletitle{Plan-on-Graph: Self-Correcting Adaptive Planning of Large Language Model on Knowledge Graphs}. In \bibinfo{booktitle}{\emph{The Thirty-eighth Annual Conference on Neural Information Processing Systems}}.
\newblock
\urldef\tempurl%
\url{https://openreview.net/forum?id=CwCUEr6wO5}
\showURL{%
\tempurl}


\bibitem[Chen et~al\mbox{.}(2023)]%
        {chen2023chatgpt}
\bibfield{author}{\bibinfo{person}{Lingjiao Chen}, \bibinfo{person}{Matei Zaharia}, {and} \bibinfo{person}{James Zou}.} \bibinfo{year}{2023}\natexlab{}.
\newblock \showarticletitle{How is ChatGPT's behavior changing over time?}
\newblock \bibinfo{journal}{\emph{arXiv preprint arXiv:2307.09009}} (\bibinfo{year}{2023}).
\newblock


\bibitem[Chen et~al\mbox{.}(2024a)]%
        {chen2024scaling}
\bibfield{author}{\bibinfo{person}{Yangyi Chen}, \bibinfo{person}{Binxuan Huang}, \bibinfo{person}{Yifan Gao}, \bibinfo{person}{Zhengyang Wang}, \bibinfo{person}{Jingfeng Yang}, {and} \bibinfo{person}{Heng Ji}.} \bibinfo{year}{2024}\natexlab{a}.
\newblock \showarticletitle{Scaling Laws for Predicting Downstream Performance in LLMs}.
\newblock \bibinfo{journal}{\emph{arXiv preprint arXiv:2410.08527}} (\bibinfo{year}{2024}).
\newblock


\bibitem[Cohere(2024)]%
        {cohere2024commandr}
\bibfield{author}{\bibinfo{person}{Cohere}.} \bibinfo{year}{2024}\natexlab{}.
\newblock \bibinfo{booktitle}{\emph{Command R: Details and Application}}.
\newblock
\urldef\tempurl%
\url{https://docs.cohere.com/v2/docs/command-r}
\showURL{%
\tempurl}
\newblock
\shownote{Accessed: 2025-02-16}.


\bibitem[Das et~al\mbox{.}(2021)]%
        {das2021case}
\bibfield{author}{\bibinfo{person}{Rajarshi Das}, \bibinfo{person}{Manzil Zaheer}, \bibinfo{person}{Dung Thai}, \bibinfo{person}{Ameya Godbole}, \bibinfo{person}{Ethan Perez}, \bibinfo{person}{Jay-Yoon Lee}, \bibinfo{person}{Lizhen Tan}, \bibinfo{person}{Lazaros Polymenakos}, {and} \bibinfo{person}{Andrew McCallum}.} \bibinfo{year}{2021}\natexlab{}.
\newblock \showarticletitle{Case-based reasoning for natural language queries over knowledge bases}.
\newblock \bibinfo{journal}{\emph{arXiv preprint arXiv:2104.08762}} (\bibinfo{year}{2021}).
\newblock


\bibitem[Ding et~al\mbox{.}(2024)]%
        {ding2024enhancing}
\bibfield{author}{\bibinfo{person}{Wentao Ding}, \bibinfo{person}{Jinmao Li}, \bibinfo{person}{Liangchuan Luo}, {and} \bibinfo{person}{Yuzhong Qu}.} \bibinfo{year}{2024}\natexlab{}.
\newblock \showarticletitle{Enhancing complex question answering over knowledge graphs through evidence pattern retrieval}. In \bibinfo{booktitle}{\emph{Proceedings of the ACM on Web Conference 2024}}. \bibinfo{pages}{2106--2115}.
\newblock


\bibitem[Dubey et~al\mbox{.}(2024)]%
        {dubey2024llama}
\bibfield{author}{\bibinfo{person}{Abhimanyu Dubey}, \bibinfo{person}{Abhinav Jauhri}, \bibinfo{person}{Abhinav Pandey}, \bibinfo{person}{Abhishek Kadian}, \bibinfo{person}{Ahmad Al-Dahle}, \bibinfo{person}{Aiesha Letman}, \bibinfo{person}{Akhil Mathur}, \bibinfo{person}{Alan Schelten}, \bibinfo{person}{Amy Yang}, \bibinfo{person}{Angela Fan}, {et~al\mbox{.}}} \bibinfo{year}{2024}\natexlab{}.
\newblock \showarticletitle{The llama 3 herd of models}.
\newblock \bibinfo{journal}{\emph{arXiv preprint arXiv:2407.21783}} (\bibinfo{year}{2024}).
\newblock


\bibitem[Dubey et~al\mbox{.}(2019)]%
        {dubey2019lc}
\bibfield{author}{\bibinfo{person}{Mohnish Dubey}, \bibinfo{person}{Debayan Banerjee}, \bibinfo{person}{Abdelrahman Abdelkawi}, {and} \bibinfo{person}{Jens Lehmann}.} \bibinfo{year}{2019}\natexlab{}.
\newblock \showarticletitle{Lc-quad 2.0: A large dataset for complex question answering over wikidata and dbpedia}. In \bibinfo{booktitle}{\emph{The Semantic Web--ISWC 2019: 18th International Semantic Web Conference, Auckland, New Zealand, October 26--30, 2019, Proceedings, Part II 18}}. Springer, \bibinfo{pages}{69--78}.
\newblock


\bibitem[Gilardi et~al\mbox{.}(2023)]%
        {gilardi2023chatgpt}
\bibfield{author}{\bibinfo{person}{Fabrizio Gilardi}, \bibinfo{person}{Meysam Alizadeh}, {and} \bibinfo{person}{Ma{\"e}l Kubli}.} \bibinfo{year}{2023}\natexlab{}.
\newblock \showarticletitle{ChatGPT outperforms crowd workers for text-annotation tasks}.
\newblock \bibinfo{journal}{\emph{Proceedings of the National Academy of Sciences}} \bibinfo{volume}{120}, \bibinfo{number}{30} (\bibinfo{year}{2023}), \bibinfo{pages}{e2305016120}.
\newblock


\bibitem[Goel et~al\mbox{.}(2020)]%
        {goel2020model}
\bibfield{author}{\bibinfo{person}{Karan Goel}, \bibinfo{person}{Albert Gu}, \bibinfo{person}{Yixuan Li}, {and} \bibinfo{person}{Christopher R{\'e}}.} \bibinfo{year}{2020}\natexlab{}.
\newblock \showarticletitle{Model patching: Closing the subgroup performance gap with data augmentation}.
\newblock \bibinfo{journal}{\emph{arXiv preprint arXiv:2008.06775}} (\bibinfo{year}{2020}).
\newblock


\bibitem[Golchin and Surdeanu(2023)]%
        {golchin2023data}
\bibfield{author}{\bibinfo{person}{Shahriar Golchin} {and} \bibinfo{person}{Mihai Surdeanu}.} \bibinfo{year}{2023}\natexlab{}.
\newblock \showarticletitle{Data contamination quiz: A tool to detect and estimate contamination in large language models}.
\newblock \bibinfo{journal}{\emph{arXiv preprint arXiv:2311.06233}} (\bibinfo{year}{2023}).
\newblock


\bibitem[Goodfellow et~al\mbox{.}(2016)]%
        {Goodfellow-et-al-2016}
\bibfield{author}{\bibinfo{person}{Ian Goodfellow}, \bibinfo{person}{Yoshua Bengio}, {and} \bibinfo{person}{Aaron Courville}.} \bibinfo{year}{2016}\natexlab{}.
\newblock \bibinfo{booktitle}{\emph{Deep Learning}}.
\newblock \bibinfo{publisher}{MIT Press}.
\newblock
\newblock
\shownote{\url{http://www.deeplearningbook.org}}.


\bibitem[Goodfellow et~al\mbox{.}(2014)]%
        {goodfellow2014explaining}
\bibfield{author}{\bibinfo{person}{Ian~J Goodfellow}, \bibinfo{person}{Jonathon Shlens}, {and} \bibinfo{person}{Christian Szegedy}.} \bibinfo{year}{2014}\natexlab{}.
\newblock \showarticletitle{Explaining and harnessing adversarial examples}.
\newblock \bibinfo{journal}{\emph{arXiv preprint arXiv:1412.6572}} (\bibinfo{year}{2014}).
\newblock


\bibitem[Gu et~al\mbox{.}(2024)]%
        {gu2024survey}
\bibfield{author}{\bibinfo{person}{Jiawei Gu}, \bibinfo{person}{Xuhui Jiang}, \bibinfo{person}{Zhichao Shi}, \bibinfo{person}{Hexiang Tan}, \bibinfo{person}{Xuehao Zhai}, \bibinfo{person}{Chengjin Xu}, \bibinfo{person}{Wei Li}, \bibinfo{person}{Yinghan Shen}, \bibinfo{person}{Shengjie Ma}, \bibinfo{person}{Honghao Liu}, {et~al\mbox{.}}} \bibinfo{year}{2024}\natexlab{}.
\newblock \showarticletitle{A Survey on LLM-as-a-Judge}.
\newblock \bibinfo{journal}{\emph{arXiv preprint arXiv:2411.15594}} (\bibinfo{year}{2024}).
\newblock


\bibitem[Gu et~al\mbox{.}(2021)]%
        {gu2021beyond}
\bibfield{author}{\bibinfo{person}{Yu Gu}, \bibinfo{person}{Sue Kase}, \bibinfo{person}{Michelle Vanni}, \bibinfo{person}{Brian Sadler}, \bibinfo{person}{Percy Liang}, \bibinfo{person}{Xifeng Yan}, {and} \bibinfo{person}{Yu Su}.} \bibinfo{year}{2021}\natexlab{}.
\newblock \showarticletitle{Beyond IID: three levels of generalization for question answering on knowledge bases}. In \bibinfo{booktitle}{\emph{Proceedings of the Web Conference 2021}}. \bibinfo{pages}{3477--3488}.
\newblock


\bibitem[Guan et~al\mbox{.}(2023)]%
        {guan2023language}
\bibfield{author}{\bibinfo{person}{Jian Guan}, \bibinfo{person}{Jesse Dodge}, \bibinfo{person}{David Wadden}, \bibinfo{person}{Minlie Huang}, {and} \bibinfo{person}{Hao Peng}.} \bibinfo{year}{2023}\natexlab{}.
\newblock \showarticletitle{Language Models Hallucinate, but May Excel at Fact Verification}.
\newblock \bibinfo{journal}{\emph{arXiv preprint arXiv:2310.14564}} (\bibinfo{year}{2023}).
\newblock


\bibitem[Guo et~al\mbox{.}(2020)]%
        {guo2020wiki}
\bibfield{author}{\bibinfo{person}{Mandy Guo}, \bibinfo{person}{Zihang Dai}, \bibinfo{person}{Denny Vrande{\v{c}}i{\'c}}, {and} \bibinfo{person}{Rami Al-Rfou}.} \bibinfo{year}{2020}\natexlab{}.
\newblock \showarticletitle{Wiki-40b: Multilingual language model dataset}. In \bibinfo{booktitle}{\emph{Proceedings of the Twelfth Language Resources and Evaluation Conference}}. \bibinfo{pages}{2440--2452}.
\newblock


\bibitem[Hoffmann et~al\mbox{.}(2022)]%
        {hoffmann2022training}
\bibfield{author}{\bibinfo{person}{Jordan Hoffmann}, \bibinfo{person}{Sebastian Borgeaud}, \bibinfo{person}{Arthur Mensch}, \bibinfo{person}{Elena Buchatskaya}, \bibinfo{person}{Trevor Cai}, \bibinfo{person}{Eliza Rutherford}, \bibinfo{person}{Diego de~Las Casas}, \bibinfo{person}{Lisa~Anne Hendricks}, \bibinfo{person}{Johannes Welbl}, \bibinfo{person}{Aidan Clark}, {et~al\mbox{.}}} \bibinfo{year}{2022}\natexlab{}.
\newblock \showarticletitle{Training compute-optimal large language models}.
\newblock \bibinfo{journal}{\emph{arXiv preprint arXiv:2203.15556}} (\bibinfo{year}{2022}).
\newblock


\bibitem[Huang et~al\mbox{.}(2023a)]%
        {huang2023chatgpt}
\bibfield{author}{\bibinfo{person}{Fan Huang}, \bibinfo{person}{Haewoon Kwak}, {and} \bibinfo{person}{Jisun An}.} \bibinfo{year}{2023}\natexlab{a}.
\newblock \showarticletitle{Is chatgpt better than human annotators? potential and limitations of chatgpt in explaining implicit hate speech}. In \bibinfo{booktitle}{\emph{Companion proceedings of the ACM web conference 2023}}. \bibinfo{pages}{294--297}.
\newblock


\bibitem[Huang et~al\mbox{.}(2023b)]%
        {huang2023survey}
\bibfield{author}{\bibinfo{person}{Lei Huang}, \bibinfo{person}{Weijiang Yu}, \bibinfo{person}{Weitao Ma}, \bibinfo{person}{Weihong Zhong}, \bibinfo{person}{Zhangyin Feng}, \bibinfo{person}{Haotian Wang}, \bibinfo{person}{Qianglong Chen}, \bibinfo{person}{Weihua Peng}, \bibinfo{person}{Xiaocheng Feng}, \bibinfo{person}{Bing Qin}, {et~al\mbox{.}}} \bibinfo{year}{2023}\natexlab{b}.
\newblock \showarticletitle{A survey on hallucination in large language models: Principles, taxonomy, challenges, and open questions}.
\newblock \bibinfo{journal}{\emph{arXiv preprint arXiv:2311.05232}} (\bibinfo{year}{2023}).
\newblock


\bibitem[Hurst et~al\mbox{.}(2024)]%
        {hurst2024gpt}
\bibfield{author}{\bibinfo{person}{Aaron Hurst}, \bibinfo{person}{Adam Lerer}, \bibinfo{person}{Adam~P Goucher}, \bibinfo{person}{Adam Perelman}, \bibinfo{person}{Aditya Ramesh}, \bibinfo{person}{Aidan Clark}, \bibinfo{person}{AJ Ostrow}, \bibinfo{person}{Akila Welihinda}, \bibinfo{person}{Alan Hayes}, \bibinfo{person}{Alec Radford}, {et~al\mbox{.}}} \bibinfo{year}{2024}\natexlab{}.
\newblock \showarticletitle{Gpt-4o system card}.
\newblock \bibinfo{journal}{\emph{arXiv preprint arXiv:2410.21276}} (\bibinfo{year}{2024}).
\newblock


\bibitem[Intelligence(2024)]%
        {Intelligence2024}
\bibfield{author}{\bibinfo{person}{Amazon Artificial~General Intelligence}.} \bibinfo{year}{2024}\natexlab{}.
\newblock \showarticletitle{The Amazon Nova family of models: Technical report and model card}.
\newblock \bibinfo{journal}{\emph{Amazon Technical Reports}} (\bibinfo{year}{2024}).
\newblock
\urldef\tempurl%
\url{https://www.amazon.science/publications/the-amazon-nova-family-of-models-technical-report-and-model-card}
\showURL{%
\tempurl}


\bibitem[Jiang et~al\mbox{.}(2023)]%
        {jiang2023reasoninglm}
\bibfield{author}{\bibinfo{person}{Jinhao Jiang}, \bibinfo{person}{Kun Zhou}, \bibinfo{person}{Wayne~Xin Zhao}, \bibinfo{person}{Yaliang Li}, {and} \bibinfo{person}{Ji-Rong Wen}.} \bibinfo{year}{2023}\natexlab{}.
\newblock \showarticletitle{Reasoninglm: Enabling structural subgraph reasoning in pre-trained language models for question answering over knowledge graph}.
\newblock \bibinfo{journal}{\emph{arXiv preprint arXiv:2401.00158}} (\bibinfo{year}{2023}).
\newblock


\bibitem[Jiang et~al\mbox{.}(2022)]%
        {jiang2022unikgqa}
\bibfield{author}{\bibinfo{person}{Jinhao Jiang}, \bibinfo{person}{Kun Zhou}, \bibinfo{person}{Wayne~Xin Zhao}, {and} \bibinfo{person}{Ji-Rong Wen}.} \bibinfo{year}{2022}\natexlab{}.
\newblock \showarticletitle{Unikgqa: Unified retrieval and reasoning for solving multi-hop question answering over knowledge graph}.
\newblock \bibinfo{journal}{\emph{arXiv preprint arXiv:2212.00959}} (\bibinfo{year}{2022}).
\newblock


\bibitem[Jin et~al\mbox{.}(2021)]%
        {jin2021disease}
\bibfield{author}{\bibinfo{person}{Di Jin}, \bibinfo{person}{Eileen Pan}, \bibinfo{person}{Nassim Oufattole}, \bibinfo{person}{Wei-Hung Weng}, \bibinfo{person}{Hanyi Fang}, {and} \bibinfo{person}{Peter Szolovits}.} \bibinfo{year}{2021}\natexlab{}.
\newblock \showarticletitle{What disease does this patient have? a large-scale open domain question answering dataset from medical exams}.
\newblock \bibinfo{journal}{\emph{Applied Sciences}} \bibinfo{volume}{11}, \bibinfo{number}{14} (\bibinfo{year}{2021}), \bibinfo{pages}{6421}.
\newblock


\bibitem[Kiela et~al\mbox{.}(2021)]%
        {kiela2021dynabench}
\bibfield{author}{\bibinfo{person}{Douwe Kiela}, \bibinfo{person}{Max Bartolo}, \bibinfo{person}{Yixin Nie}, \bibinfo{person}{Divyansh Kaushik}, \bibinfo{person}{Atticus Geiger}, \bibinfo{person}{Zhengxuan Wu}, \bibinfo{person}{Bertie Vidgen}, \bibinfo{person}{Grusha Prasad}, \bibinfo{person}{Amanpreet Singh}, \bibinfo{person}{Pratik Ringshia}, {et~al\mbox{.}}} \bibinfo{year}{2021}\natexlab{}.
\newblock \showarticletitle{Dynabench: Rethinking benchmarking in NLP}.
\newblock \bibinfo{journal}{\emph{arXiv preprint arXiv:2104.14337}} (\bibinfo{year}{2021}).
\newblock


\bibitem[Kojima et~al\mbox{.}(2022)]%
        {kojima2022large}
\bibfield{author}{\bibinfo{person}{Takeshi Kojima}, \bibinfo{person}{Shixiang~Shane Gu}, \bibinfo{person}{Machel Reid}, \bibinfo{person}{Yutaka Matsuo}, {and} \bibinfo{person}{Yusuke Iwasawa}.} \bibinfo{year}{2022}\natexlab{}.
\newblock \showarticletitle{Large language models are zero-shot reasoners}.
\newblock \bibinfo{journal}{\emph{Advances in neural information processing systems}}  \bibinfo{volume}{35} (\bibinfo{year}{2022}), \bibinfo{pages}{22199--22213}.
\newblock


\bibitem[Lan and Jiang(2020)]%
        {lan2020query}
\bibfield{author}{\bibinfo{person}{Yunshi Lan} {and} \bibinfo{person}{Jing Jiang}.} \bibinfo{year}{2020}\natexlab{}.
\newblock \showarticletitle{Query graph generation for answering multi-hop complex questions from knowledge bases}. Association for Computational Linguistics.
\newblock


\bibitem[Lehmann et~al\mbox{.}(2015)]%
        {lehmann2015dbpedia}
\bibfield{author}{\bibinfo{person}{Jens Lehmann}, \bibinfo{person}{Robert Isele}, \bibinfo{person}{Max Jakob}, \bibinfo{person}{Anja Jentzsch}, \bibinfo{person}{Dimitris Kontokostas}, \bibinfo{person}{Pablo~N Mendes}, \bibinfo{person}{Sebastian Hellmann}, \bibinfo{person}{Mohamed Morsey}, \bibinfo{person}{Patrick Van~Kleef}, \bibinfo{person}{S{\"o}ren Auer}, {et~al\mbox{.}}} \bibinfo{year}{2015}\natexlab{}.
\newblock \showarticletitle{Dbpedia--a large-scale, multilingual knowledge base extracted from wikipedia}.
\newblock \bibinfo{journal}{\emph{Semantic web}} \bibinfo{volume}{6}, \bibinfo{number}{2} (\bibinfo{year}{2015}), \bibinfo{pages}{167--195}.
\newblock


\bibitem[Lewis(2019)]%
        {lewis2019bart}
\bibfield{author}{\bibinfo{person}{Mike Lewis}.} \bibinfo{year}{2019}\natexlab{}.
\newblock \showarticletitle{Bart: Denoising sequence-to-sequence pre-training for natural language generation, translation, and comprehension}.
\newblock \bibinfo{journal}{\emph{arXiv preprint arXiv:1910.13461}} (\bibinfo{year}{2019}).
\newblock


\bibitem[Lewis et~al\mbox{.}(2020)]%
        {lewis2020retrieval}
\bibfield{author}{\bibinfo{person}{Patrick Lewis}, \bibinfo{person}{Ethan Perez}, \bibinfo{person}{Aleksandra Piktus}, \bibinfo{person}{Fabio Petroni}, \bibinfo{person}{Vladimir Karpukhin}, \bibinfo{person}{Naman Goyal}, \bibinfo{person}{Heinrich K{\"u}ttler}, \bibinfo{person}{Mike Lewis}, \bibinfo{person}{Wen-tau Yih}, \bibinfo{person}{Tim Rockt{\"a}schel}, {et~al\mbox{.}}} \bibinfo{year}{2020}\natexlab{}.
\newblock \showarticletitle{Retrieval-augmented generation for knowledge-intensive nlp tasks}.
\newblock \bibinfo{journal}{\emph{Advances in Neural Information Processing Systems}}  \bibinfo{volume}{33} (\bibinfo{year}{2020}), \bibinfo{pages}{9459--9474}.
\newblock


\bibitem[Li et~al\mbox{.}(2023)]%
        {Li2023HaluEvalAL}
\bibfield{author}{\bibinfo{person}{Junyi Li}, \bibinfo{person}{Xiaoxue Cheng}, \bibinfo{person}{Wayne~Xin Zhao}, \bibinfo{person}{Jianyun Nie}, {and} \bibinfo{person}{Ji rong Wen}.} \bibinfo{year}{2023}\natexlab{}.
\newblock \showarticletitle{HaluEval: A Large-Scale Hallucination Evaluation Benchmark for Large Language Models}.
\newblock \bibinfo{journal}{\emph{ArXiv}}  \bibinfo{volume}{abs/2305.11747} (\bibinfo{year}{2023}).
\newblock
\urldef\tempurl%
\url{https://api.semanticscholar.org/CorpusID:258832847}
\showURL{%
\tempurl}


\bibitem[Li(2023)]%
        {li2023open}
\bibfield{author}{\bibinfo{person}{Yucheng Li}.} \bibinfo{year}{2023}\natexlab{}.
\newblock \showarticletitle{An open source data contamination report for llama series models}.
\newblock \bibinfo{journal}{\emph{arXiv preprint arXiv:2310.17589}} (\bibinfo{year}{2023}).
\newblock


\bibitem[Li et~al\mbox{.}(2024)]%
        {li2024survey}
\bibfield{author}{\bibinfo{person}{Yongqi Li}, \bibinfo{person}{Xinyu Lin}, \bibinfo{person}{Wenjie Wang}, \bibinfo{person}{Fuli Feng}, \bibinfo{person}{Liang Pang}, \bibinfo{person}{Wenjie Li}, \bibinfo{person}{Liqiang Nie}, \bibinfo{person}{Xiangnan He}, {and} \bibinfo{person}{Tat-Seng Chua}.} \bibinfo{year}{2024}\natexlab{}.
\newblock \showarticletitle{A survey of generative search and recommendation in the era of large language models}.
\newblock \bibinfo{journal}{\emph{arXiv preprint arXiv:2404.16924}} (\bibinfo{year}{2024}).
\newblock


\bibitem[LUO et~al\mbox{.}(2024)]%
        {luo2024reasoning}
\bibfield{author}{\bibinfo{person}{LINHAO LUO}, \bibinfo{person}{Yuan-Fang Li}, \bibinfo{person}{Reza Haf}, {and} \bibinfo{person}{Shirui Pan}.} \bibinfo{year}{2024}\natexlab{}.
\newblock \showarticletitle{Reasoning on Graphs: Faithful and Interpretable Large Language Model Reasoning}. In \bibinfo{booktitle}{\emph{The Twelfth International Conference on Learning Representations}}.
\newblock
\urldef\tempurl%
\url{https://openreview.net/forum?id=ZGNWW7xZ6Q}
\showURL{%
\tempurl}


\bibitem[Ma et~al\mbox{.}(2021)]%
        {ma2021dynaboard}
\bibfield{author}{\bibinfo{person}{Zhiyi Ma}, \bibinfo{person}{Kawin Ethayarajh}, \bibinfo{person}{Tristan Thrush}, \bibinfo{person}{Somya Jain}, \bibinfo{person}{Ledell Wu}, \bibinfo{person}{Robin Jia}, \bibinfo{person}{Christopher Potts}, \bibinfo{person}{Adina Williams}, {and} \bibinfo{person}{Douwe Kiela}.} \bibinfo{year}{2021}\natexlab{}.
\newblock \showarticletitle{Dynaboard: An evaluation-as-a-service platform for holistic next-generation benchmarking}.
\newblock \bibinfo{journal}{\emph{Advances in Neural Information Processing Systems}}  \bibinfo{volume}{34} (\bibinfo{year}{2021}), \bibinfo{pages}{10351--10367}.
\newblock


\bibitem[Mehlhorn(1988)]%
        {mehlhorn1988faster}
\bibfield{author}{\bibinfo{person}{Kurt Mehlhorn}.} \bibinfo{year}{1988}\natexlab{}.
\newblock \showarticletitle{A faster approximation algorithm for the Steiner problem in graphs}.
\newblock \bibinfo{journal}{\emph{Inform. Process. Lett.}} \bibinfo{volume}{27}, \bibinfo{number}{3} (\bibinfo{year}{1988}), \bibinfo{pages}{125--128}.
\newblock


\bibitem[Miller et~al\mbox{.}(2016)]%
        {miller2016key}
\bibfield{author}{\bibinfo{person}{Alexander Miller}, \bibinfo{person}{Adam Fisch}, \bibinfo{person}{Jesse Dodge}, \bibinfo{person}{Amir-Hossein Karimi}, \bibinfo{person}{Antoine Bordes}, {and} \bibinfo{person}{Jason Weston}.} \bibinfo{year}{2016}\natexlab{}.
\newblock \showarticletitle{Key-value memory networks for directly reading documents}.
\newblock \bibinfo{journal}{\emph{arXiv preprint arXiv:1606.03126}} (\bibinfo{year}{2016}).
\newblock


\bibitem[Min et~al\mbox{.}(2023)]%
        {min2023factscore}
\bibfield{author}{\bibinfo{person}{Sewon Min}, \bibinfo{person}{Kalpesh Krishna}, \bibinfo{person}{Xinxi Lyu}, \bibinfo{person}{Mike Lewis}, \bibinfo{person}{Wen tau Yih}, \bibinfo{person}{Pang~Wei Koh}, \bibinfo{person}{Mohit Iyyer}, \bibinfo{person}{Luke Zettlemoyer}, {and} \bibinfo{person}{Hannaneh Hajishirzi}.} \bibinfo{year}{2023}\natexlab{}.
\newblock \bibinfo{title}{FActScore: Fine-grained Atomic Evaluation of Factual Precision in Long Form Text Generation}.
\newblock
\newblock
\showeprint[arxiv]{2305.14251}~[cs.CL]


\bibitem[Morris et~al\mbox{.}(2020)]%
        {morris2020textattack}
\bibfield{author}{\bibinfo{person}{John~X Morris}, \bibinfo{person}{Eli Lifland}, \bibinfo{person}{Jin~Yong Yoo}, \bibinfo{person}{Jake Grigsby}, \bibinfo{person}{Di Jin}, {and} \bibinfo{person}{Yanjun Qi}.} \bibinfo{year}{2020}\natexlab{}.
\newblock \showarticletitle{Textattack: A framework for adversarial attacks, data augmentation, and adversarial training in nlp}.
\newblock \bibinfo{journal}{\emph{arXiv preprint arXiv:2005.05909}} (\bibinfo{year}{2020}).
\newblock


\bibitem[Ngomo(2018)]%
        {ngomo20189th}
\bibfield{author}{\bibinfo{person}{Ngonga Ngomo}.} \bibinfo{year}{2018}\natexlab{}.
\newblock \showarticletitle{9th challenge on question answering over linked data (QALD-9)}.
\newblock \bibinfo{journal}{\emph{language}} \bibinfo{volume}{7}, \bibinfo{number}{1} (\bibinfo{year}{2018}), \bibinfo{pages}{58--64}.
\newblock


\bibitem[Ostyakova et~al\mbox{.}(2023)]%
        {ostyakova2023chatgpt}
\bibfield{author}{\bibinfo{person}{Lidiia Ostyakova}, \bibinfo{person}{Veronika Smilga}, \bibinfo{person}{Kseniia Petukhova}, \bibinfo{person}{Maria Molchanova}, {and} \bibinfo{person}{Daniel Kornev}.} \bibinfo{year}{2023}\natexlab{}.
\newblock \showarticletitle{Chatgpt vs. crowdsourcing vs. experts: Annotating open-domain conversations with speech functions}. In \bibinfo{booktitle}{\emph{Proceedings of the 24th Annual Meeting of the Special Interest Group on Discourse and Dialogue}}. \bibinfo{pages}{242--254}.
\newblock


\bibitem[Pellissier~Tanon et~al\mbox{.}(2016)]%
        {pellissier2016freebase}
\bibfield{author}{\bibinfo{person}{Thomas Pellissier~Tanon}, \bibinfo{person}{Denny Vrande{\v{c}}i{\'c}}, \bibinfo{person}{Sebastian Schaffert}, \bibinfo{person}{Thomas Steiner}, {and} \bibinfo{person}{Lydia Pintscher}.} \bibinfo{year}{2016}\natexlab{}.
\newblock \showarticletitle{From freebase to wikidata: The great migration}. In \bibinfo{booktitle}{\emph{Proceedings of the 25th international conference on world wide web}}. \bibinfo{pages}{1419--1428}.
\newblock


\bibitem[Raffel et~al\mbox{.}(2020)]%
        {raffel2020exploring}
\bibfield{author}{\bibinfo{person}{Colin Raffel}, \bibinfo{person}{Noam Shazeer}, \bibinfo{person}{Adam Roberts}, \bibinfo{person}{Katherine Lee}, \bibinfo{person}{Sharan Narang}, \bibinfo{person}{Michael Matena}, \bibinfo{person}{Yanqi Zhou}, \bibinfo{person}{Wei Li}, {and} \bibinfo{person}{Peter~J Liu}.} \bibinfo{year}{2020}\natexlab{}.
\newblock \showarticletitle{Exploring the limits of transfer learning with a unified text-to-text transformer}.
\newblock \bibinfo{journal}{\emph{Journal of machine learning research}} \bibinfo{volume}{21}, \bibinfo{number}{140} (\bibinfo{year}{2020}), \bibinfo{pages}{1--67}.
\newblock


\bibitem[Rawles et~al\mbox{.}(2024)]%
        {rawles2024androidworld}
\bibfield{author}{\bibinfo{person}{Christopher Rawles}, \bibinfo{person}{Sarah Clinckemaillie}, \bibinfo{person}{Yifan Chang}, \bibinfo{person}{Jonathan Waltz}, \bibinfo{person}{Gabrielle Lau}, \bibinfo{person}{Marybeth Fair}, \bibinfo{person}{Alice Li}, \bibinfo{person}{William Bishop}, \bibinfo{person}{Wei Li}, \bibinfo{person}{Folawiyo Campbell-Ajala}, {et~al\mbox{.}}} \bibinfo{year}{2024}\natexlab{}.
\newblock \showarticletitle{AndroidWorld: A dynamic benchmarking environment for autonomous agents}.
\newblock \bibinfo{journal}{\emph{arXiv preprint arXiv:2405.14573}} (\bibinfo{year}{2024}).
\newblock


\bibitem[Saxena et~al\mbox{.}(2020)]%
        {saxena2020improving}
\bibfield{author}{\bibinfo{person}{Apoorv Saxena}, \bibinfo{person}{Aditay Tripathi}, {and} \bibinfo{person}{Partha Talukdar}.} \bibinfo{year}{2020}\natexlab{}.
\newblock \showarticletitle{Improving multi-hop question answering over knowledge graphs using knowledge base embeddings}. In \bibinfo{booktitle}{\emph{Proceedings of the 58th annual meeting of the association for computational linguistics}}. \bibinfo{pages}{4498--4507}.
\newblock


\bibitem[Schoenegger et~al\mbox{.}(2024)]%
        {schoenegger2024wisdom}
\bibfield{author}{\bibinfo{person}{Philipp Schoenegger}, \bibinfo{person}{Indre Tuminauskaite}, \bibinfo{person}{Peter~S Park}, {and} \bibinfo{person}{Philip~E Tetlock}.} \bibinfo{year}{2024}\natexlab{}.
\newblock \showarticletitle{Wisdom of the silicon crowd: Llm ensemble prediction capabilities match human crowd accuracy}.
\newblock \bibinfo{journal}{\emph{arXiv preprint arXiv:2402.19379}} (\bibinfo{year}{2024}).
\newblock


\bibitem[Suchanek et~al\mbox{.}(2024)]%
        {suchanek2024yago}
\bibfield{author}{\bibinfo{person}{Fabian~M Suchanek}, \bibinfo{person}{Mehwish Alam}, \bibinfo{person}{Thomas Bonald}, \bibinfo{person}{Lihu Chen}, \bibinfo{person}{Pierre-Henri Paris}, {and} \bibinfo{person}{Jules Soria}.} \bibinfo{year}{2024}\natexlab{}.
\newblock \showarticletitle{Yago 4.5: A large and clean knowledge base with a rich taxonomy}. In \bibinfo{booktitle}{\emph{Proceedings of the 47th International ACM SIGIR Conference on Research and Development in Information Retrieval}}. \bibinfo{pages}{131--140}.
\newblock


\bibitem[Sun et~al\mbox{.}(2019)]%
        {sun2019pullnet}
\bibfield{author}{\bibinfo{person}{Haitian Sun}, \bibinfo{person}{Tania Bedrax-Weiss}, {and} \bibinfo{person}{William~W Cohen}.} \bibinfo{year}{2019}\natexlab{}.
\newblock \showarticletitle{Pullnet: Open domain question answering with iterative retrieval on knowledge bases and text}.
\newblock \bibinfo{journal}{\emph{arXiv preprint arXiv:1904.09537}} (\bibinfo{year}{2019}).
\newblock


\bibitem[Sun et~al\mbox{.}(2018)]%
        {sun2018open}
\bibfield{author}{\bibinfo{person}{Haitian Sun}, \bibinfo{person}{Bhuwan Dhingra}, \bibinfo{person}{Manzil Zaheer}, \bibinfo{person}{Kathryn Mazaitis}, \bibinfo{person}{Ruslan Salakhutdinov}, {and} \bibinfo{person}{William~W Cohen}.} \bibinfo{year}{2018}\natexlab{}.
\newblock \showarticletitle{Open domain question answering using early fusion of knowledge bases and text}.
\newblock \bibinfo{journal}{\emph{arXiv preprint arXiv:1809.00782}} (\bibinfo{year}{2018}).
\newblock


\bibitem[Sun et~al\mbox{.}(2024)]%
        {sun2024thinkongraph}
\bibfield{author}{\bibinfo{person}{Jiashuo Sun}, \bibinfo{person}{Chengjin Xu}, \bibinfo{person}{Lumingyuan Tang}, \bibinfo{person}{Saizhuo Wang}, \bibinfo{person}{Chen Lin}, \bibinfo{person}{Yeyun Gong}, \bibinfo{person}{Lionel Ni}, \bibinfo{person}{Heung-Yeung Shum}, {and} \bibinfo{person}{Jian Guo}.} \bibinfo{year}{2024}\natexlab{}.
\newblock \showarticletitle{Think-on-Graph: Deep and Responsible Reasoning of Large Language Model on Knowledge Graph}. In \bibinfo{booktitle}{\emph{The Twelfth International Conference on Learning Representations}}.
\newblock
\urldef\tempurl%
\url{https://openreview.net/forum?id=nnVO1PvbTv}
\showURL{%
\tempurl}


\bibitem[Talmor and Berant(2018)]%
        {talmor2018web}
\bibfield{author}{\bibinfo{person}{Alon Talmor} {and} \bibinfo{person}{Jonathan Berant}.} \bibinfo{year}{2018}\natexlab{}.
\newblock \showarticletitle{The web as a knowledge-base for answering complex questions}.
\newblock \bibinfo{journal}{\emph{arXiv preprint arXiv:1803.06643}} (\bibinfo{year}{2018}).
\newblock


\bibitem[Team(2025)]%
        {mistral2025small3}
\bibfield{author}{\bibinfo{person}{Mistral~AI Team}.} \bibinfo{year}{2025}\natexlab{}.
\newblock \bibinfo{booktitle}{\emph{Mistral Small 3}}.
\newblock
\urldef\tempurl%
\url{https://mistral.ai/en/news/mistral-small-3}
\showURL{%
\tempurl}
\newblock
\shownote{Accessed: 2025-02-16}.


\bibitem[Trivedi et~al\mbox{.}(2017)]%
        {trivedi2017lc}
\bibfield{author}{\bibinfo{person}{Priyansh Trivedi}, \bibinfo{person}{Gaurav Maheshwari}, \bibinfo{person}{Mohnish Dubey}, {and} \bibinfo{person}{Jens Lehmann}.} \bibinfo{year}{2017}\natexlab{}.
\newblock \showarticletitle{Lc-quad: A corpus for complex question answering over knowledge graphs}. In \bibinfo{booktitle}{\emph{The Semantic Web--ISWC 2017: 16th International Semantic Web Conference, Vienna, Austria, October 21-25, 2017, Proceedings, Part II 16}}. Springer, \bibinfo{pages}{210--218}.
\newblock


\bibitem[Usbeck et~al\mbox{.}(2024)]%
        {usbeck2024qald}
\bibfield{author}{\bibinfo{person}{Ricardo Usbeck}, \bibinfo{person}{Xi Yan}, \bibinfo{person}{Aleksandr Perevalov}, \bibinfo{person}{Longquan Jiang}, \bibinfo{person}{Julius Schulz}, \bibinfo{person}{Angelie Kraft}, \bibinfo{person}{Cedric M{\"o}ller}, \bibinfo{person}{Junbo Huang}, \bibinfo{person}{Jan Reineke}, \bibinfo{person}{Axel-Cyrille Ngonga~Ngomo}, {et~al\mbox{.}}} \bibinfo{year}{2024}\natexlab{}.
\newblock \showarticletitle{QALD-10--The 10th challenge on question answering over linked data: Shifting from DBpedia to Wikidata as a KG for KGQA}.
\newblock \bibinfo{journal}{\emph{Semantic Web}} \bibinfo{volume}{15}, \bibinfo{number}{6} (\bibinfo{year}{2024}), \bibinfo{pages}{2193--2207}.
\newblock


\bibitem[Vrande{\v{c}}i{\'c} and Kr{\"o}tzsch(2014)]%
        {vrandevcic2014wikidata}
\bibfield{author}{\bibinfo{person}{Denny Vrande{\v{c}}i{\'c}} {and} \bibinfo{person}{Markus Kr{\"o}tzsch}.} \bibinfo{year}{2014}\natexlab{}.
\newblock \showarticletitle{Wikidata: a free collaborative knowledgebase}.
\newblock \bibinfo{journal}{\emph{Commun. ACM}} \bibinfo{volume}{57}, \bibinfo{number}{10} (\bibinfo{year}{2014}), \bibinfo{pages}{78--85}.
\newblock


\bibitem[Vukov et~al\mbox{.}(2023)]%
        {vukov2023ouroboros}
\bibfield{author}{\bibinfo{person}{Joseph~Michael Vukov}, \bibinfo{person}{Tera~Lynn Joseph}, \bibinfo{person}{Gina Lebkuecher}, \bibinfo{person}{Michelle Ramirez}, {and} \bibinfo{person}{Michael~B Burns}.} \bibinfo{year}{2023}\natexlab{}.
\newblock \showarticletitle{The Ouroboros Threat}.
\newblock \bibinfo{journal}{\emph{The American Journal of Bioethics}} \bibinfo{volume}{23}, \bibinfo{number}{10} (\bibinfo{year}{2023}), \bibinfo{pages}{58--60}.
\newblock


\bibitem[Wei et~al\mbox{.}(2022)]%
        {wei2022chain}
\bibfield{author}{\bibinfo{person}{Jason Wei}, \bibinfo{person}{Xuezhi Wang}, \bibinfo{person}{Dale Schuurmans}, \bibinfo{person}{Maarten Bosma}, \bibinfo{person}{Fei Xia}, \bibinfo{person}{Ed Chi}, \bibinfo{person}{Quoc~V Le}, \bibinfo{person}{Denny Zhou}, {et~al\mbox{.}}} \bibinfo{year}{2022}\natexlab{}.
\newblock \showarticletitle{Chain-of-thought prompting elicits reasoning in large language models}.
\newblock \bibinfo{journal}{\emph{Advances in neural information processing systems}}  \bibinfo{volume}{35} (\bibinfo{year}{2022}), \bibinfo{pages}{24824--24837}.
\newblock


\bibitem[Wu et~al\mbox{.}(2023)]%
        {wu2023retrieve}
\bibfield{author}{\bibinfo{person}{Yike Wu}, \bibinfo{person}{Nan Hu}, \bibinfo{person}{Sheng Bi}, \bibinfo{person}{Guilin Qi}, \bibinfo{person}{Jie Ren}, \bibinfo{person}{Anhuan Xie}, {and} \bibinfo{person}{Wei Song}.} \bibinfo{year}{2023}\natexlab{}.
\newblock \showarticletitle{Retrieve-rewrite-answer: A kg-to-text enhanced llms framework for knowledge graph question answering}.
\newblock \bibinfo{journal}{\emph{arXiv preprint arXiv:2309.11206}} (\bibinfo{year}{2023}).
\newblock


\bibitem[Xu et~al\mbox{.}(2024)]%
        {xu2024hallucinationinevitableinnatelimitation}
\bibfield{author}{\bibinfo{person}{Ziwei Xu}, \bibinfo{person}{Sanjay Jain}, {and} \bibinfo{person}{Mohan Kankanhalli}.} \bibinfo{year}{2024}\natexlab{}.
\newblock \bibinfo{title}{Hallucination is Inevitable: An Innate Limitation of Large Language Models}.
\newblock
\newblock
\showeprint[arxiv]{2401.11817}~[cs.CL]
\urldef\tempurl%
\url{https://arxiv.org/abs/2401.11817}
\showURL{%
\tempurl}


\bibitem[Yang et~al\mbox{.}(2023)]%
        {yang2023rethinking}
\bibfield{author}{\bibinfo{person}{Shuo Yang}, \bibinfo{person}{Wei-Lin Chiang}, \bibinfo{person}{Lianmin Zheng}, \bibinfo{person}{Joseph~E Gonzalez}, {and} \bibinfo{person}{Ion Stoica}.} \bibinfo{year}{2023}\natexlab{}.
\newblock \showarticletitle{Rethinking benchmark and contamination for language models with rephrased samples}.
\newblock \bibinfo{journal}{\emph{arXiv preprint arXiv:2311.04850}} (\bibinfo{year}{2023}).
\newblock


\bibitem[Yao et~al\mbox{.}(2024)]%
        {yao2024tau}
\bibfield{author}{\bibinfo{person}{Shunyu Yao}, \bibinfo{person}{Noah Shinn}, \bibinfo{person}{Pedram Razavi}, {and} \bibinfo{person}{Karthik Narasimhan}.} \bibinfo{year}{2024}\natexlab{}.
\newblock \showarticletitle{$tau$-bench: A Benchmark for Tool-Agent-User Interaction in Real-World Domains}.
\newblock \bibinfo{journal}{\emph{arXiv preprint arXiv:2406.12045}} (\bibinfo{year}{2024}).
\newblock


\bibitem[Yih et~al\mbox{.}(2016)]%
        {yih2016value}
\bibfield{author}{\bibinfo{person}{Wen-tau Yih}, \bibinfo{person}{Matthew Richardson}, \bibinfo{person}{Christopher Meek}, \bibinfo{person}{Ming-Wei Chang}, {and} \bibinfo{person}{Jina Suh}.} \bibinfo{year}{2016}\natexlab{}.
\newblock \showarticletitle{The value of semantic parse labeling for knowledge base question answering}. In \bibinfo{booktitle}{\emph{Proceedings of the 54th Annual Meeting of the Association for Computational Linguistics (Volume 2: Short Papers)}}. \bibinfo{pages}{201--206}.
\newblock


\bibitem[Zhang et~al\mbox{.}(2022)]%
        {zhang2022subgraph}
\bibfield{author}{\bibinfo{person}{Jing Zhang}, \bibinfo{person}{Xiaokang Zhang}, \bibinfo{person}{Jifan Yu}, \bibinfo{person}{Jian Tang}, \bibinfo{person}{Jie Tang}, \bibinfo{person}{Cuiping Li}, {and} \bibinfo{person}{Hong Chen}.} \bibinfo{year}{2022}\natexlab{}.
\newblock \showarticletitle{Subgraph retrieval enhanced model for multi-hop knowledge base question answering}.
\newblock \bibinfo{journal}{\emph{arXiv preprint arXiv:2202.13296}} (\bibinfo{year}{2022}).
\newblock


\bibitem[Zhang et~al\mbox{.}(2018)]%
        {zhang2018variational}
\bibfield{author}{\bibinfo{person}{Yuyu Zhang}, \bibinfo{person}{Hanjun Dai}, \bibinfo{person}{Zornitsa Kozareva}, \bibinfo{person}{Alexander Smola}, {and} \bibinfo{person}{Le Song}.} \bibinfo{year}{2018}\natexlab{}.
\newblock \showarticletitle{Variational reasoning for question answering with knowledge graph}. In \bibinfo{booktitle}{\emph{Proceedings of the AAAI conference on artificial intelligence}}, Vol.~\bibinfo{volume}{32}.
\newblock


\bibitem[Zhang et~al\mbox{.}(2023)]%
        {zhang2023siren}
\bibfield{author}{\bibinfo{person}{Yue Zhang}, \bibinfo{person}{Yafu Li}, \bibinfo{person}{Leyang Cui}, \bibinfo{person}{Deng Cai}, \bibinfo{person}{Lemao Liu}, \bibinfo{person}{Tingchen Fu}, \bibinfo{person}{Xinting Huang}, \bibinfo{person}{Enbo Zhao}, \bibinfo{person}{Yu Zhang}, \bibinfo{person}{Yulong Chen}, {et~al\mbox{.}}} \bibinfo{year}{2023}\natexlab{}.
\newblock \showarticletitle{Siren's Song in the AI Ocean: A Survey on Hallucination in Large Language Models}.
\newblock \bibinfo{journal}{\emph{arXiv preprint arXiv:2309.01219}} (\bibinfo{year}{2023}).
\newblock


\bibitem[Zhang et~al\mbox{.}(2024)]%
        {zhang2024darg}
\bibfield{author}{\bibinfo{person}{Zhehao Zhang}, \bibinfo{person}{Jiaao Chen}, {and} \bibinfo{person}{Diyi Yang}.} \bibinfo{year}{2024}\natexlab{}.
\newblock \showarticletitle{Darg: Dynamic evaluation of large language models via adaptive reasoning graph}.
\newblock \bibinfo{journal}{\emph{arXiv preprint arXiv:2406.17271}} (\bibinfo{year}{2024}).
\newblock


\bibitem[Zheng et~al\mbox{.}(2024)]%
        {zheng2024judging}
\bibfield{author}{\bibinfo{person}{Lianmin Zheng}, \bibinfo{person}{Wei-Lin Chiang}, \bibinfo{person}{Ying Sheng}, \bibinfo{person}{Siyuan Zhuang}, \bibinfo{person}{Zhanghao Wu}, \bibinfo{person}{Yonghao Zhuang}, \bibinfo{person}{Zi Lin}, \bibinfo{person}{Zhuohan Li}, \bibinfo{person}{Dacheng Li}, \bibinfo{person}{Eric Xing}, {et~al\mbox{.}}} \bibinfo{year}{2024}\natexlab{}.
\newblock \showarticletitle{Judging llm-as-a-judge with mt-bench and chatbot arena}.
\newblock \bibinfo{journal}{\emph{Advances in Neural Information Processing Systems}}  \bibinfo{volume}{36} (\bibinfo{year}{2024}).
\newblock


\bibitem[Zhou et~al\mbox{.}(2023)]%
        {zhou2023don}
\bibfield{author}{\bibinfo{person}{Kun Zhou}, \bibinfo{person}{Yutao Zhu}, \bibinfo{person}{Zhipeng Chen}, \bibinfo{person}{Wentong Chen}, \bibinfo{person}{Wayne~Xin Zhao}, \bibinfo{person}{Xu Chen}, \bibinfo{person}{Yankai Lin}, \bibinfo{person}{Ji-Rong Wen}, {and} \bibinfo{person}{Jiawei Han}.} \bibinfo{year}{2023}\natexlab{}.
\newblock \showarticletitle{Don't make your llm an evaluation benchmark cheater}.
\newblock \bibinfo{journal}{\emph{arXiv preprint arXiv:2311.01964}} (\bibinfo{year}{2023}).
\newblock


\bibitem[Zhu et~al\mbox{.}(2023)]%
        {Zhu2023DyValDE}
\bibfield{author}{\bibinfo{person}{Kaijie Zhu}, \bibinfo{person}{Jiaao Chen}, \bibinfo{person}{Jindong Wang}, \bibinfo{person}{Neil~Zhenqiang Gong}, \bibinfo{person}{Diyi Yang}, {and} \bibinfo{person}{Xing Xie}.} \bibinfo{year}{2023}\natexlab{}.
\newblock \showarticletitle{DyVal: Dynamic Evaluation of Large Language Models for Reasoning Tasks}. In \bibinfo{booktitle}{\emph{International Conference on Learning Representations}}.
\newblock
\urldef\tempurl%
\url{https://api.semanticscholar.org/CorpusID:263310319}
\showURL{%
\tempurl}


\bibitem[Zhu et~al\mbox{.}(2024)]%
        {Zhu2024DyVal2D}
\bibfield{author}{\bibinfo{person}{Kaijie Zhu}, \bibinfo{person}{Jindong Wang}, \bibinfo{person}{Qinlin Zhao}, \bibinfo{person}{Ruochen Xu}, {and} \bibinfo{person}{Xing Xie}.} \bibinfo{year}{2024}\natexlab{}.
\newblock \showarticletitle{DyVal 2: Dynamic Evaluation of Large Language Models by Meta Probing Agents}.
\newblock \bibinfo{journal}{\emph{ArXiv}}  \bibinfo{volume}{abs/2402.14865} (\bibinfo{year}{2024}).
\newblock
\urldef\tempurl%
\url{https://api.semanticscholar.org/CorpusID:267897463}
\showURL{%
\tempurl}


\bibitem[Zong et~al\mbox{.}(2023)]%
        {zong2023fool}
\bibfield{author}{\bibinfo{person}{Yongshuo Zong}, \bibinfo{person}{Tingyang Yu}, \bibinfo{person}{Ruchika Chavhan}, \bibinfo{person}{Bingchen Zhao}, {and} \bibinfo{person}{Timothy Hospedales}.} \bibinfo{year}{2023}\natexlab{}.
\newblock \showarticletitle{Fool your (vision and) language model with embarrassingly simple permutations}.
\newblock \bibinfo{journal}{\emph{arXiv preprint arXiv:2310.01651}} (\bibinfo{year}{2023}).
\newblock


\end{thebibliography}

\clearpage

\section*{Appendix}
\label{appendix}

\subsection*{Prompts}

Here, we provide the prompts used in the \textsc{Dynamic-KGQA} framework for QA generation (P1) and evaluation (P2, P3), as shown in Figure~\ref{fig:pipeline}.

\label{app:prompts}

\begin{tcolorbox}[
    colback=lightgray!20, 
    colframe=gray, 
    title={\textbf{Prompt P1: Knowledge Graph Q\&A Generation}}, 
    fonttitle=\bfseries,
    sharp corners=southwest,
    coltitle=black,
    label=prompt:p1
]

\small 

You are an AI assistant tasked with generating question-answer pairs from knowledge graph triples. Your goal is to create natural, human-like questions and their corresponding answers based on the provided graph data.

\textbf{Task Overview:} \newline
Generate \textbf{multi-hop, complex Q\&A pairs} where the questions appear simple and natural but require reasoning across multiple connected relationships within the graph to infer the answer.

\textbf{Guidelines for Generating Q\&A Pairs:}
\begin{enumerate}
    \item \textbf{Question Design:}
    \begin{itemize}
        \item Questions should utilize multiple connected relationships in the graph, requiring multi-hop reasoning.
        \item Avoid single-hop or trivial questions directly derived from a single triple.
        \item The answer should be an entity or node in the graph.
    \end{itemize}
    
    \item \textbf{Multi-Hop Reasoning:}
    \begin{itemize}
        \item Use paths connecting entities indirectly through multiple relationships to infer answers.
        \item Questions should reflect meaningful and interesting connections within the graph.
        \item Aim for questions with at least 4 hops or higher whenever possible.
    \end{itemize}
    
    \item \textbf{Answer Validation:}
    \begin{itemize}
        \item Ensure each answer is fully supported by one or more triples from the graph.
        \item Include the exact path (a sequence of triples) that justifies the answer.
    \end{itemize}
    
    \item \textbf{Comprehensive Coverage:}
    \begin{itemize}
        \item Generate as many high-quality Q\&A pairs as possible, exploring all meaningful paths and connections within the graph.
    \end{itemize}
    
    \item \textbf{Fallback Condition:}
    \begin{itemize}
        \item If no valid Q\&A pairs can be generated from the graph, explicitly indicate this in the response.
    \end{itemize}
\end{enumerate}

\vspace{0.01cm}
\textbf{Graph Data:} \newline
Below is the graph data for your task: \texttt{\{triples\_str\}}

\vspace{0.01cm}
\textbf{Respond ONLY with JSON in the following structure:}
\begin{itemize}
    \item \texttt{valid\_qa\_pairs}: Boolean indicating if valid QA pairs were generated.
    \item \texttt{number\_of\_qa\_pairs}: Integer specifying the total number of QA pairs.
    \item \texttt{qa\_pairs}: A list of QA pairs, where each pair includes:
    \begin{itemize}
        \item \texttt{question}: String representing the question.
        \item \texttt{answer}: String representing the answer.
        \item \texttt{supporting\_path}: A list of triples, where each triple includes:
        \begin{itemize}
            \item \texttt{subject}: String representing the subject.
            \item \texttt{predicate}: String representing predicate.
            \item \texttt{object}: String representing the object.
        \end{itemize}
    \end{itemize}
\end{itemize}

\end{tcolorbox}

\begin{tcolorbox}[
    colback=lightgray!20, 
    colframe=gray, 
    title={\textbf{Prompt P2: Question Evaluation Prompt}}, 
    fonttitle=\bfseries,
    sharp corners=southwest,
    coltitle=black,
    label=prompt:p2
]
\small

As an expert evaluator, your role is to assess the quality and validity of trivia or natural questions. These questions aim to test the responder's knowledge, which may require implicit or external information. Your goal is to analyze the question based on the following criteria:

\textbf{Logical Structure}: Verify if the grammar and syntax are correct. (\textbf{True} if grammatically and syntactically correct; \textbf{False} if there are issues with grammar or syntax.)

\textbf{Redundancy}: Confirm that the question does not contain its own answer explicitly or through overly obvious phrasing. (\textbf{True} if it contains its answer; \textbf{False} if it does not.)

\vspace{0.01cm}
\textbf{Output JSON Keys:}
\begin{itemize}
    \item \texttt{question}: The input question.
    \item \texttt{logical\_structure\_flag}: (\textbf{True/False})
    \item \texttt{logical\_structure\_reasoning}: Reason for the logical structure flag.
    \item \texttt{redundancy\_flag}: (\textbf{True/False})
    \item \texttt{redundancy\_reasoning}: Reason for redundancy flag.
\end{itemize}

\vspace{0.01cm}
\textbf{Task:} \newline
Analyze the following question and provide a JSON object containing flags and reasons for potential issues:

\textbf{Question}: \texttt{\{question\}}

\vspace{0.01cm}
\textbf{Output:} \newline
Return a JSON object that evaluates the question based on the criteria above.
\end{tcolorbox}

\begin{tcolorbox}[
    colback=lightgray!20, 
    colframe=gray, 
    title={\textbf{Prompt P3: Answer Evaluation Prompt}}, 
    fonttitle=\bfseries,
    sharp corners=southwest,
    coltitle=black,
    label=prompt:p3
]
\small 

As an expert evaluator, your role is to assess the validity and adequacy of an answer to a given question based on a provided set of supporting facts (triples). The goal is to determine if the answer is logically supported by the facts and if it sufficiently addresses the question. Your evaluation should consider the following criteria:

\textbf{Answer Support}: Verify if the answer is explicitly or implicitly supported by the provided supporting facts. (\textbf{True} if the supporting facts justify the answer; \textbf{False} if they do not.)

\textbf{Answer Adequacy}: Determine if the answer fully and clearly addresses the question. An adequate answer should provide a direct and comprehensive response to the question. (\textbf{True} if the answer is adequate; \textbf{False} if the answer is incomplete or does not directly address the question.)

\vspace{0.01cm}
\textbf{Output JSON Keys:}
\begin{itemize}
    \item \texttt{answer\_support\_flag}: (\textbf{True/False})
    \item \texttt{answer\_support\_reasoning}: Reason for the answer support flag.
    \item \texttt{answer\_adequacy\_flag}: (\textbf{True/False})
    \item \texttt{answer\_adequacy\_reasoning}: Reason for the answer adequacy flag.
\end{itemize}

\vspace{0.01cm}
\textbf{Inputs:}
\begin{itemize}
    \item \textbf{Question}: \texttt{\{question\}}
    \item \textbf{Answer}: \texttt{\{answer\}}
    \item \textbf{Supporting Facts}: \texttt{\{supporting\_facts\}}
\end{itemize}

\vspace{0.01cm}
\textbf{Task:} \newline
Analyze the question, answer, and supporting facts. Return a JSON object containing flags and reasoning based on the criteria above.

\vspace{0.01cm}
\textbf{Output:} \newline
Return only the JSON object based on the criteria above.

\end{tcolorbox}

\end{document}